\documentclass[style=google]{fairmeta}
\usepackage{mathptmx}
\usepackage{helvet}     
\usepackage{courier} 
\usepackage{xcolor}
\usepackage{enumitem}
\usepackage{float}
\usepackage{xltabular}
\usepackage{pgfplots}
\pgfplotsset{compat=1.18}
\usepackage{wrapfig}
\usepackage{svg}
\usepackage{tcolorbox}
\usepackage{hyperref}
\usepackage{url}
\usepackage{multirow}
\usepackage{multicol}
\usepackage{booktabs}
\usepackage{graphicx} 
\usepackage{adjustbox}
\usepackage{colortbl}   
\usepackage{makecell}
\usepackage[utf8]{inputenc}
\usepackage{subcaption}
\usepackage{epstopdf}
\usepackage{longtable}
\usepackage{listings}
\usepackage{verbatim}
\usepackage{array}
\usepackage{textcomp}
\usepackage{nicefrac} 
\usepackage{placeins}

\usepackage{devanagari}

\definecolor{lightgray}{gray}{0.95}
\lstdefinestyle{chatml}{
    backgroundcolor=\color{lightgray},
    basicstyle=\ttfamily,
    frame=none,
    columns=fullflexible,
    breaklines=true,
    keepspaces=true,
    showstringspaces=false
}

\title{\textsc{BhashaBench V1}: A Comprehensive Benchmark for the Quadrant of Indic Domains}

\author{Vijay Devane}
\author{Mohd Nauman}
\author{Bhargav Patel}
\author{Aniket Wakchoure}
\author{Yogeshkumar Sant}
\author{Shyam Pawar}
\author{Viraj Thakur}
\author{Ananya Godse}
\author{Sunil Patra}
\author{Neha Maurya}
\author{Suraj Racha}
\author{Nitish Singh}
\author{Ajay Nagpal}
\author{Piyush Sawarkar}
\author{Kundeshwar Pundalik}
\author{Rohit Saluja}
\author{Ganesh Ramakrishnan}

\affiliation{BharatGen Team}

\abstract{
The rapid advancement of large language models (LLMs) has intensified the need for domain and culture specific evaluation. Existing benchmarks are largely Anglocentric and domain-agnostic, limiting their applicability to India-centric contexts. To address this gap, we introduce \textbf{BhashaBench V1}, the first domain-specific, multi-task, bilingual benchmark focusing on critical Indic knowledge systems. BhashaBench V1 contains \textbf{74,166} meticulously curated question-answer pairs, with 52,494 in English and 21,672 in Hindi, sourced from authentic government and domain-specific exams. It spans four major domains: Agriculture, Legal, Finance, and Ayurveda, comprising 90+ subdomains and covering 500+ topics, enabling fine-grained evaluation. Evaluation of 29+ LLMs reveals significant domain and language specific performance gaps, with especially large disparities in low-resource domains. For instance, GPT-4o achieves 76.49\% overall accuracy in Legal but only 59.74\% in Ayurveda. Models consistently perform better on English content compared to Hindi across all domains. Subdomain-level analysis shows that areas such as \textit{Cyber Law}, \textit{International Finance} perform relatively well, while \textit{Panchakarma}, \textit{Seed Science}, and \textit{Human Rights} remain notably weak. \textbf{BhashaBench V1} provides a comprehensive dataset for evaluating large language models across India's diverse knowledge domains. It enables assessment of models' ability to integrate domain-specific knowledge with bilingual understanding. All code, benchmarks, and resources are publicly available to support open research: \url{https://bharatgen-iitb-tih.github.io/bhashabenchv1/}
}

\date{\today}
\correspondence{Kundeshwar.pundalik@tihiitb.org }

\begin{document}
\maketitle

\section{Introduction}
The rapid advancement of large language models (LLMs) has transformed artificial intelligence, extending their capabilities far beyond traditional natural language processing. Models such as GPT-4o \citep{GPT4o}, GPT-OSS-120B \citep{openai2025gptoss120bgptoss20bmodel}, DeepSeek-V3 \citep{deepseekai2025deepseekv3technicalreport}, and Qwen-3 \citep{yang2025qwen3technicalreport} excel across diverse domains, from code generation and mathematical reasoning to creative writing and scientific analysis \citep{brown2020language, touvron2023llama2openfoundation, openai2024gpt4technicalreport}, enabling applications in conversational AI, education, healthcare, finance, legal services, and agriculture \citep{bubeck2023sparks, wei2022emergent}. Platforms like \textit{Krishi Sathi \citep{Krishisathi2025}} leverage LLMs for crop advisory and pest detection, improving agricultural productivity. Despite these advances, substantial performance gaps remain in multilingual and domain-specific contexts, particularly for non-Latin, low-resource languages \citep{wang2024multilingualpretrainingusinglarge, zhong2025opportunitieschallengeslargelanguage, ahuja2024megaversebenchmarkinglargelanguage}. English-centric training limits models’ ability to capture nuanced knowledge in specialized fields and India-specific domains, such as Ayurveda, indigenous agriculture, finance, and regional legal systems \citep{winata2021languagemodelsfewshotmultilingual, sen2023indic, khanuja2021murilmultilingualrepresentationsindian}, highlighting the need for culturally and contextually aware evaluation.
\begin{figure}[t]
    \centering
    \resizebox{\textwidth}{0.5\textheight}{
    \includegraphics{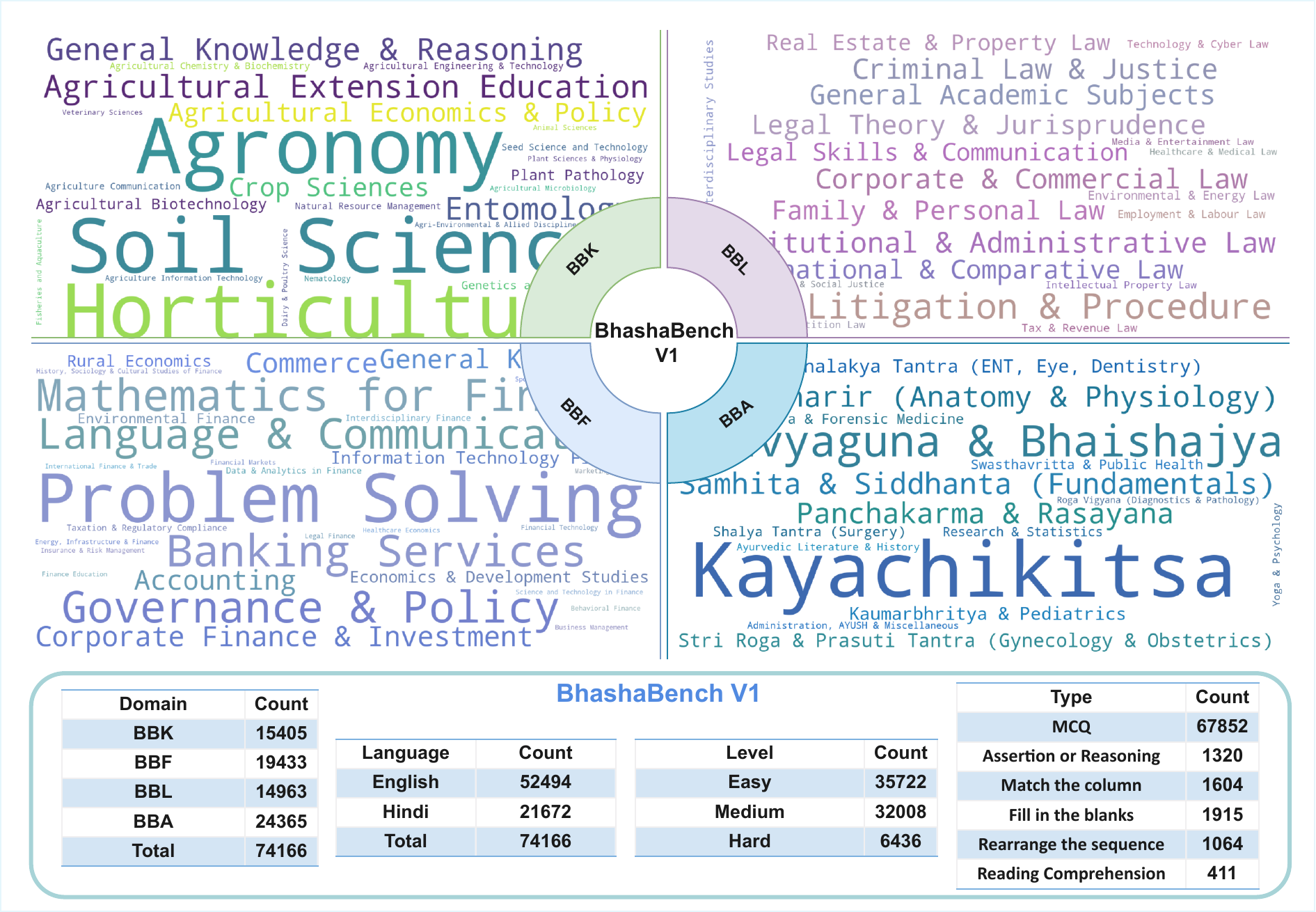}
    }
\vspace{-0.6cm}
    \caption{Overview diagram and statistics of BhashaBench V1.}
    \label{fig:bhasha_bench_v1_main}
\end{figure}
The scale of this problem demands urgent attention, as India's diverse knowledge ecosystem affects millions of lives across multiple critical domains. In agriculture alone, Over 40 million farmers rely on farming-related activities \citep{indiadatamap2025}, and access to accurate information on crop management, soil health, and sustainable practices can have a direct impact on food security and livelihoods. The complexity is further magnified by the fact that each state in India has its own distinct agricultural methods, crop varieties, soil conditions, and traditional farming practices that have evolved over centuries to suit local climatic and geographical conditions. Similarly, India's legal system processes millions of cases annually, requiring precise understanding of complex legal frameworks, precedents, and procedural nuances that vary across states and jurisdictions \citep{njdg2023statistics}. The healthcare sector, particularly traditional medicine systems like Ayurveda, serves millions of patients who rely on practitioners' knowledge of ancient texts, formulations, and treatment protocols. Furthermore, India's financial ecosystem processes billions of transactions daily, including over 100 billion UPI transactions annually, where even minor misunderstandings in financial regulations or procedures can have cascading effects \citep{npci2023statistics}. While existing benchmarks such as MMLU \citep{hendrycks2021measuringmassivemultitasklanguage}, HellaSwag \citep{zellers2019hellaswagmachinereallyfinish}, AGIEVAL \citep{zhong2023agieval}, and more recent multilingual efforts like MEGA \citep{ahuja2024megaversebenchmarkinglargelanguage} attempt to assess model capabilities, they often focus primarily on English content and may not fully capture India-specific nuances, cultural contexts, and domain expertise that are essential for real-world applications in the Indian subcontinent.

To address these critical gaps, we introduce \textbf{BhashaBench V1}, the first comprehensive domain-specific, multi-task, bilingual benchmark designed explicitly for evaluating large language models on India-centric knowledge systems. BhashaBench V1 encompasses four fundamental domains that form the backbone of Indian society and economy: Agriculture (BBK - BhashaBench Krishi), Legal (BBL - BhashaBench Legal), Finance (BBF - BhashaBench Finance), and Ayurveda (BBA - BhashaBench Ayurveda). The benchmark spans over 90 subdomains and covers more than 500 specific topics, reflecting the intricate complexity and diversity of Indian knowledge systems. This granular categorization enables fine-grained evaluation of model performance across specialized areas that require deep domain expertise and cultural understanding. The dataset has been meticulously curated from over 40 authentic government and professional examination papers, ensuring that the questions reflect real-world scenarios and ground-level challenges faced by practitioners in these domains \citep{iks2025, ZHONG2024105316}. To maximize coverage across India's linguistic landscape, BhashaBench V1 currently supports English and Hindi, the two most widely understood languages in the country, collectively enabling assessment of models' capabilities for a significant portion of India's population while maintaining the cultural and contextual authenticity of the original knowledge systems.

Our comprehensive evaluation of 29+ state-of-the-art language models on BhashaBench V1 reveals significant performance disparities across domains and languages, highlighting the urgent need for India-specific model development and evaluation. The results demonstrate substantial domain-specific performance gaps, with models showing varying degrees of competency across different knowledge areas. For instance, GPT-4o, one of the top-performing models, achieved 76.49\% accuracy in the Legal domain but only 59.74\% in Ayurveda, illustrating the challenges models face with traditional Indian knowledge systems. Similarly, consistent language-specific performance gaps emerged, with models generally performing better on English content compared to Hindi across all domains. The subdomain-level analysis further reveals granular insights into model capabilities, showing that certain areas such as Cyber Law and International Finance demonstrate relatively strong performance, while traditional domains like Panchakarma, Seed Science, and Human Rights remain notably challenging for current LLMs. These findings underscore the critical importance of domain and language-specific evaluation frameworks for assessing model readiness for real-world deployment in diverse Indian contexts.

\section{ RELATED WORK}
\label{related work}
\subsection{Exploration of LLMs}
The landscape of large language models has witnessed unprecedented growth, with both proprietary and open-source models achieving remarkable capabilities. Recent proprietary LLMs, including GPT-4o and GPT-4o-mini \citep{openai2024gpt4technicalreport}, Claude-3.5 Sonnet \citep{anthropic2024claude3}, and the Gemini series \citep{google2023gemini}, have demonstrated significant improvements across various benchmarks \citep{chiang2024chatbot, wang2024language}. The open-source ecosystem has flourished with models such as the Llama-3 series \citep{grattafiori2024llama3herdmodels}, Gemma \citep{gemmateam2024gemmaopenmodelsbased}, Qwen2.5 \citep{yang2025qwen}, and Mistral \citep{jiang2023mistral7b} achieving competitive performance while maintaining transparency and accessibility.

While primarily trained on English-dominant corpora, many models incorporate substantial multilingual data during pretraining \citep{gemmateam2024gemmaopenmodelsbased, grattafiori2024llama3herdmodels, ustun2024ayamodelinstructionfinetuned}, enabling capabilities in hundreds of languages with varying proficiency \citep{nguyen2023seallm}. Language-specific models have gained momentum, particularly for underrepresented languages including Indic languages \citep{gala2024airavata, gala2023indictrans2highqualityaccessiblemachine}. Notable examples include Airavata \citep{gala2024airavata}, MuRIL \citep{khanuja2021muril}, and recent generative models like Param-1 \citep{pundalik2025param1bharatgen29bmodel}.

Domain-specific language models have emerged as a critical research direction. Medical applications include Med-PaLM \citep{singhal2023large} and BioBERT \citep{Lee_2019}, while legal and financial domains have seen LegalBERT \citep{chalkidis2020legal} and FinBERT \citep{yang2020finbert} respectively. In the Indian context, domain-specific initiatives like Agri-Param \citep{bharatgenai_agriparam_2025}, Ayur-Param \citep{bharatgenai_ayurparam_2025}, Finance-Param \citep{bharatgenai_financeparam_2025}, and Legal-Param \citep{bharatgenai_legalparam_2025} address unique requirements of India's diverse knowledge systems through continual pretraining \citep{nag2024efficientcontinualpretrainingllms} or instruction fine-tuning \citep{aralimatti2025finetuningsmalllanguagemodels}.

Despite these advances, comprehensive evaluation frameworks for culturally and linguistically diverse domains remain limited, particularly for traditional knowledge systems requiring nuanced understanding of local contexts. This work conducts a comprehensive evaluation of 29+ state-of-the-art models on BhashaBench V1 to address these evaluation challenges.

\subsection{Evaluation of LLMs}
Numerous benchmarks have been developed to assess large language model performance. General-purpose benchmarks such as MMLU \citep{hendrycks2021mmlu}, MMLU-Pro \citep{wang2024mmlu-pro}, AGIEval \citep{zhong2023agieval}, BIG-Bench \citep{srivastava2023imitationgamequantifyingextrapolating}, and HellaSwag   \citep{zellers2019hellaswagmachinereallyfinish} evaluate LLMs across diverse tasks from commonsense reasoning to knowledge-intensive question answering. However, these remain largely Anglocentric with limited multilingual evaluation \citep{Bandarkar_2024, kakwani-etal-2020-indicnlpsuite}.

To address domain-specific challenges, specialized benchmarks have emerged. In agriculture, benchmarks like AgriBench \citep{zhou2024agribench}, BVL QA Corpus \citep{bvl2024}, AgXQA \citep{msu2024agxqa}, AgEval \citep{rauf2024ageval}, and SeedBench \citep{seedbench2025} cover crop disease identification to advisory support. The finance domain features FinGAIA \citep{finGAIA2025}, FinanceBench \citep{financebench2023}, MultiFin \citep{multifinben2025}, InvestorBench \citep{investorbench2024}, and MultiFinBen \citep{multifinben2025} for financial reasoning, fraud detection, and trading evaluation. Legal domain efforts include IL-TUR \citep{iltur2024}, IndicLegalQA \citep{indicalqa2024}, LegalBench \citep{guha2023legalbench}, LEXTREME \citep{chalkidis2023lextreme}, and the CAIL series \citep{xiao2018cail, xiao2019cail} for legal question answering, case summarization, and judgment prediction. Traditional medicine resources such as MTCMB \citep{mtcmb2025}, Pratyaya-Kosh \citep{ragad2019ayurvedic}, Anveshana \citep{terdalkar2023ayurjnanam}, and OpenTCM \citep{opentcm2025} provide task-specific evaluation datasets covering knowledge graphs, OCR correction, and dosha analysis.

Despite this progress, key limitations persist. Many benchmarks are restricted to English or high-resource languages, limiting effectiveness for multilingual and Indic contexts. Others focus on narrow tasks, unable to capture full domain expertise. Evaluation methodologies vary widely from accuracy scores to human judgments, hindering standardized comparison across domains and languages. These gaps underscore the need for a unified, multilingual, and domain-aware evaluation framework.

\section{BhashaBench V1}

\subsection{Design Principles}

The primary motivation behind BhashaBench V1 is to comprehensively assess domain-specific knowledge and reasoning capabilities of large language models within India's diverse and culturally rich knowledge ecosystems. Unlike existing benchmarks focusing on general or Western-centric domains, our benchmark evaluates specialized Indian knowledge systems requiring deep cultural understanding and contextual awareness. BhashaBench V1 adheres to seven core design principles: \textbf{(1) Critical Indian Domains:} Encompasses Agriculture, Legal systems, Finance, and Ayurveda with fine-grained subfields. \textbf{(2) Diverse Task Formats:} Includes multiple-choice, assertion-reasoning, fill-in-blanks, and comprehension tasks. \textbf{(3) India-Specific Reasoning:} Evaluates domain-specific reasoning incorporating cultural contexts and regional practices. \textbf{(4) Bilingual Framework:} Supports English and Hindi evaluation maintaining cultural authenticity. \textbf{(5) Authentic Sources:} Questions curated from government examinations and professional certifications. \textbf{(6) Difficulty Assessment:} Categorized into Easy, Medium, Hard levels. \textbf{(7) Cultural Authenticity:} Prioritizes traditional knowledge systems including Ayurvedic principles. \footnote{More collection and processing procedures can be found in Appendix \ref{app:dataset}.} This framework spans 90+ subdomains covering 500+ topics, enabling comprehensive evaluation of model capabilities in India-centric contexts.

\subsection{Data Collection}

The data collection process for BhashaBench V1 follows a systematic approach similar to AGIEVAL \citep{zhong2023agieval}, focusing on authentic examination materials from national and state-level assessments. We systematically gathered publicly available question papers from official online examination portals, which host previously released papers that are manually curated by subject matter experts, ensuring accurate topic tagging, language annotation, and validated answer keys.

Our comprehensive collection encompasses over 40 different examination types across multiple categories: national competitive exams, domain-specific degree examinations, professional certification tests, and state-level civil services examinations. Regional state examinations proved particularly valuable as they incorporate state-specific topics, local knowledge systems, and cultural practices often overlooked in national assessments. These examinations are typically taken by individuals seeking higher education opportunities or career advancement, ensuring questions reflect practical, real-world knowledge requirements.

The final dataset comprises \textbf{74,166} carefully curated question–answer pairs spanning four core domains, with \textbf{52,494 questions in English} (70.8\%) and \textbf{21,672 questions in Hindi} (29.2\%), reflecting practical usage patterns in Indian educational and professional contexts. This approach ensures BhashaBench V1 captures the nuanced intersection between language, culture, and domain expertise essential for effective model deployment in Indian contexts.

\subsection{Data Processing}

Our data processing phase focused on extracting structured question-answer pairs from PDF examination papers while preserving linguistic and formatting nuances essential for authentic evaluation. Most examination materials were available exclusively in PDF format, requiring sophisticated OCR processing pipelines to handle multilingual content and domain-specific terminology.

\textbf{OCR Pipeline Selection:} Based on existing evaluations \citep{vikram2024surya}, Surya OCR demonstrated superior performance in handling Indic languages and domain-specific content. Reported results show 98.1\% normalized text similarity for English and 98.9\% for Hindi, with an average of 97.8\%, outperforming alternatives such as Tesseract (88.0\% overall) and Google Vision API (96.7\%). Surya's architecture, designed for multilingual document understanding with enhanced Indic script support, makes it a suitable choice for diverse examination materials.

\textbf{Question-Answer Extraction Pipeline:} Following OCR processing, we developed an extraction pipeline leveraging GPT-OSS-120B \citep{gpt2024oss} to structure raw text into formatted question-answer pairs. Key challenges included format variations across examination bodies, answer key alignment, multi-format questions (MCQ, assertion-reasoning, comprehension), and language-specific formatting conventions. The pipeline included: (1) \textbf{Question Extraction} using GPT-OSS-120B for boundary detection across different layouts; (2) \textbf{Option Parsing} to maintain original labeling conventions; (3) \textbf{Answer Key Alignment} processing both inline and separate answer documents; and (4) \textbf{Format Standardization} into consistent JSON structure with domain metadata.

\textbf{Data Cleaning and Quality Control:} Our multi-layered cleaning approach addressed noise and inconsistencies through systematic filtering. We excluded image-based questions, and questions with more than four options. Language verification used INDICLID \citep{madhani2023indiclid} and Unicode-based filtering \citep{Khan_2024} for proper linguistic categorization. Approximately 30\% of questions lacked subdomain classification, addressed using GPT-OSS-120B with domain-specific taxonomies. We classified questions into six categories: MCQ, assertion-reasoning, fill-in-the-blanks, match-the-column, reading comprehension, and sequence rearrangement. Duplicate detection employed both exact-match and semantic similarity measures.

\textbf{Manual Validation:} Following a methodology similar to \citep{Bandarkar_2024}, all extracted question-answer pairs underwent rigorous expert validation to ensure accuracy verification, cultural context preservation, ambiguity resolution, and consistency standardization. Additionally, domain experts reviewed the linguistic authenticity to maintain the natural flow and idiomatic expressions characteristic of each language. This comprehensive multi-stage validation approach ensured that BhashaBench V1 maintains the highest data quality standards while preserving the authentic complexity and cultural specificity of the original examination materials.

\subsection{Data Analysis}
\begin{figure}[!t]
  \centering
   \begin{minipage}{0.46\textwidth}
   \centering
   \resizebox{1.\textwidth}{!}{
   \includegraphics{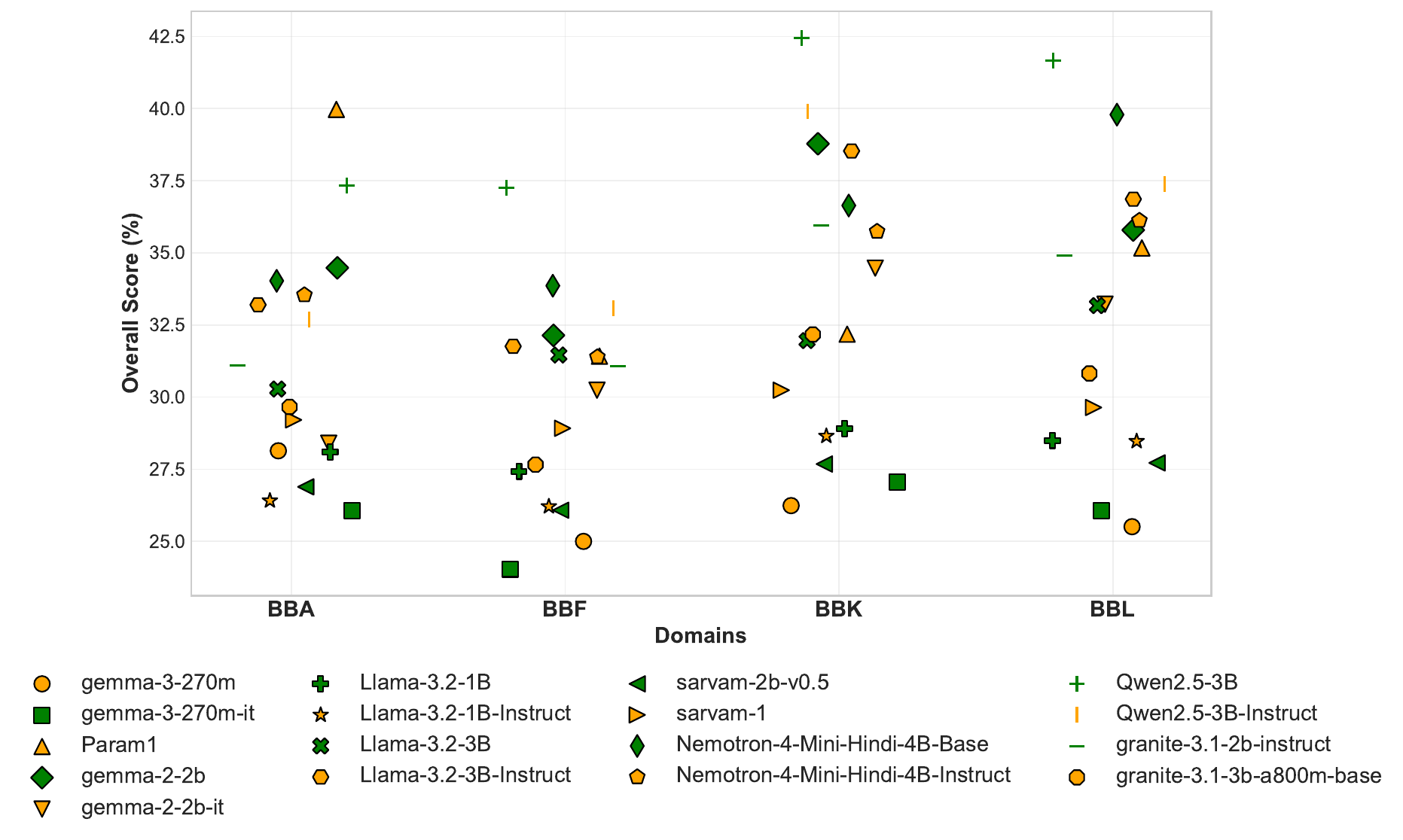}
   }
   \vspace{-0.6cm}
   \caption{Comparative performance of small models ($\leq$4B) over BhashaBench V1.}
   \label{fig:model_performance_domains}
 \end{minipage}
   \hfill
  \begin{minipage}{0.52\textwidth}
   \centering
    \resizebox{1.\textwidth}{!}{
      \includegraphics{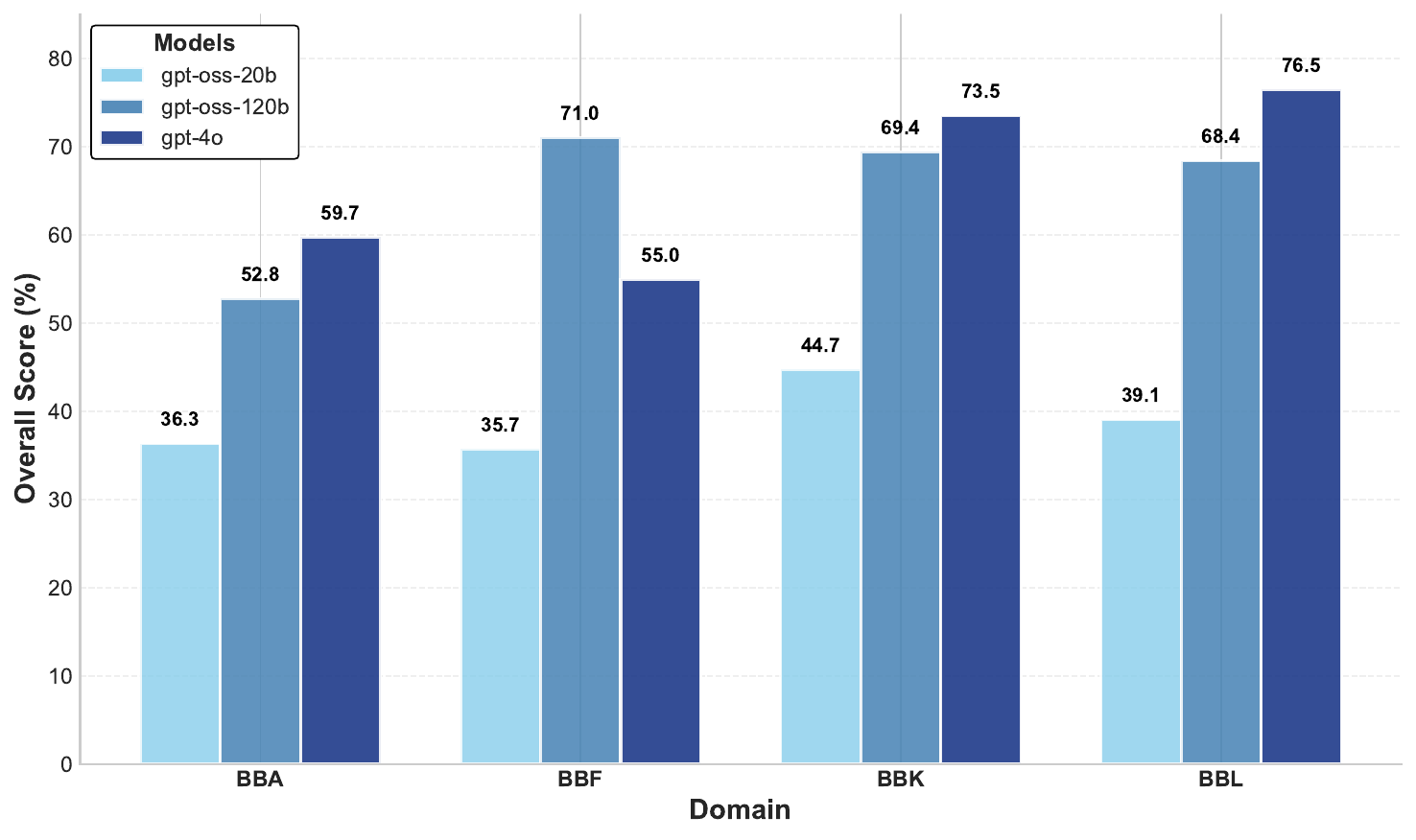}
      }
\vspace{-0.6cm}
  \caption{Comparative performance analysis of the GPT model family on BhashaBench V1.}
  \label{fig:model_performance_comparison}
  \end{minipage}
 \end{figure}

Figure~\ref{fig:bhasha_bench_v1_main} presents the comprehensive statistics of BhashaBench V1. Detailed exposition is provided in Appendix~\ref{subapp:data_analysis}. Of the total 74,166 questions, 70.8\% are in English while 29.2\% are in Hindi, reflecting practical bilingual usage patterns in Indian professional contexts. The dataset spans four specialized domains with varying complexity levels across 91 subdomains.

\textbf{Agriculture (BBK):} This domain encompasses agricultural sciences relevant to Indian farming systems across 25 subdomains. Agronomy dominates with 5,078 questions, reflecting its foundational role in agricultural education. The domain covers traditional practices alongside emerging areas like Agricultural Biotechnology and IT solutions. Its balanced difficulty distribution (44\% easy, 45\% medium, 11\% hard) ensures comprehensive skill assessment.

\textbf{Finance (BBF):} Covers India's complex financial ecosystem through 30 subdomains. Problem Solving leads with 5,686 questions, followed by Mathematics for Finance (4,845), emphasizing the quantitative nature of financial practice. The domain uniquely incorporates India-specific areas like Rural Economics and Environmental Finance while addressing modern fintech developments.

\textbf{Ayurveda (BBA):} Represents traditional Indian medicine across 16 subdomains. Kayachikitsa (General Medicine) forms the core with 3,134 questions, while Dravyaguna covers pharmacology and therapeutics (2,972). This domain shows the highest proportion of accessible questions (53\% easy), reflecting its foundational knowledge structure.

\textbf{Legal (BBL):} Encompasses Indian jurisprudence through 20 subdomains. Civil Litigation \& Procedure dominates with 7,126 questions, followed by Constitutional Law (3,609). The domain balances traditional legal areas with contemporary developments like Technology \& Cyber Law, maintaining strong cultural relevance through Family \& Personal Law.

The predominantly MCQ format ($>$90\%) ensures consistent evaluation methodology while supporting diverse cognitive assessment approaches across India-specific knowledge systems.

\section{Experimental Setup}

We evaluate multiple state-of-the-art models on BhashaBench V1, including large proprietary models, open-source multilingual models, and domain-specific fine-tuned variants. Both base versions and instruction fine-tuned models are assessed to measure the effectiveness of specialized training approaches across India-specific knowledge domains. All evaluations are conducted in a zero-shot setting to assess the models' inherent capabilities without task-specific examples. For open-source models, we utilize the LM-EVALUATION-HARNESS library~\citep{gao2024lmevaluation, biderman2024lmevaluation} to ensure clean, reproducible, and standardized evaluations. We employ the log-likelihood method where the probability of a given output string is computed by conditioning it on the provided input~\citep{brown2020language}. For multiple choice questions with $k$ possible answer choices, we select the answer string $(a_i)$ with the highest conditional log probability: $\arg\max(\log P(a_1|x), ..., \log P(a_k|x))$.

For closed-source and large-scale proprietary models, we utilize their respective APIs for evaluation due to computational constraints and access limitations. These API-based models are evaluated using a generative approach and are prompted to generate responses in a structured JSON format to facilitate automated response parsing. This comprehensive experimental framework enables systematic comparison across diverse model architectures while maintaining evaluation consistency across both open-source and proprietary systems. Additional details regarding model specifications, hyperparameters, and computational resources are provided in Appendix~\ref{app:experimental_setup}.

\section{Results and Discussions}
In this section, we discuss the results and our findings across all the experiments conducted.
\subsection{Zero-Shot Performance Across All Domains (EN + HI)}

Table~\ref{tab:results_discussions} shows the performance of various models in English and Hindi under the zero-shot setup. Among these, Qwen3-235B-A22B-Instruct emerges as the strongest model, consistently outperforming all competitors across both languages, with an average accuracy of 67.25\%. This is followed by GPT-4O at 66.18\% and gpt-oss-120b at 65.41\%. Performance shows clear stratification across model sizes and types, with models exceeding 27B parameters demonstrating substantially higher accuracies compared to smaller variants. Among the 7B-27B range, gemma-2-27b leads with 53.11\% average accuracy, followed by gemma-2-27b-it at 44.64\%. In the mid-range category, gemma-2-9b shows impressive performance at 48.07\%, with Pangea-7B achieving 41.54\%.

Smaller models under 4B parameters show more modest performance, with Qwen2.5-3B achieving the highest accuracy in this category at 39.68\%. Models specifically designed for Indian languages include Param-1 (34.69\%) and the Nemotron-4-Mini-Hindi variants (36.08\% and 34.20\%). Performance is notably higher in English compared to Hindi across most models, reflecting the typical pattern observed in multilingual language models, with models showing varying degrees of cross-lingual transfer capabilities.
\begin{table}[!t]
    \centering
    \caption{Zero-shot scores (\%) of LLMs across domains on BhashaBench V1 (EN + HI). The benchmark covers Agriculture (BBK), Finance (BBF), Legal (BBL), and Ayurveda (BBA). ``Avg'' denotes the overall average across that domain.}

    \resizebox{\textwidth}{!}{
    \begin{tabular}{c|ccc|ccc|ccc|ccc}
    \toprule
    \multirow{2}{*}{Model} & \multicolumn{3}{c|}{BBA} & \multicolumn{3}{c|}{BBF} & \multicolumn{3}{c|}{BBK} & \multicolumn{3}{c}{BBL} \\ 
    \cmidrule{2-13}
    & Eng & Hin & Avg & Eng & Hin & Avg & Eng & Hin & Avg & Eng & Hin & Avg \\ 
    \midrule
    \multicolumn{13}{c}{\textit{\textless\ 4B Models}} \\
    \midrule
    gemma-3-270m & 28.08 & 28.25 & 28.14 & 24.98 & 25.06 & 25.00 & 26.64 & 24.45 & 26.24 & 25.49 & 25.54 & 25.51 \\
    gemma-3-270m-it & 26.23 & 25.77 & 26.06 & 24.13 & 23.84 & 24.04 & 27.44 & 25.35 & 27.06 & 25.56 & 27.26 & 26.07 \\
    Param-1 & 41.12 & 38.04 & 39.97 & 32.24 & 29.56 & 31.42 & 33.10 & 27.97 & 32.18 & 36.15 & 32.89 & 35.17 \\
    gemma-2-2b & 36.80 & 30.61 & 34.48 & 34.20 & 27.50 & 32.14 & 41.24 & 27.49 & 38.78 & 38.45 & 29.61 & 35.79 \\
    gemma-2-2b-it & 29.38 & 26.79 & 28.40 & 31.26 & 27.93 & 30.24 & 35.94 & 27.71 & 34.47 & 34.49 & 30.25 & 33.22 \\
    Llama-3.2-1B & 29.17 & 26.30 & 28.10 & 28.24 & 25.61 & 27.43 & 29.71 & 25.21 & 28.91 & 29.63 & 25.88 & 28.52 \\
    Llama-3.2-1B-Instruct & 26.77 & 25.82 & 26.41 & 26.28 & 26.04 & 26.21 & 29.16 & 26.33 & 28.65 & 29.08 & 27.04 & 28.47 \\
    Llama-3.2-3B & 31.62 & 28.05 & 30.28 & 33.04 & 27.92 & 31.46 & 32.68 & 28.69 & 31.96 & 35.17 & 28.53 & 33.17 \\
    Llama-3.2-3B-Instruct & 35.31 & 29.67 & 33.20 & 32.94 & 29.09 & 31.76 & 40.59 & 29.09 & 38.53 & 39.74 & 30.13 & 36.86 \\
    sarvam-2b-v0.5 & 26.79 & 27.07 & 26.89 & 26.42 & 25.31 & 26.08 & 28.14 & 25.57 & 27.68 & 28.49 & 25.95 & 27.72 \\
    sarvam-1 & 29.70 & 28.41 & 29.21 & 29.66 & 27.27 & 28.92 & 30.82 & 27.57 & 30.24 & 30.92 & 26.66 & 29.64 \\
    Nemotron-4-Mini-Hindi-4B-Base & 34.76 & 32.82 & 34.03 & 34.95 & 31.41 & 33.86 & 36.67 & 36.49 & 36.64 & 40.75 & 37.55 & 39.79 \\
    Nemotron-4-Mini-Hindi-4B-Instruct & 33.38 & 33.82 & 33.54 & 31.98 & 30.06 & 31.39 & 35.83 & 35.33 & 35.74 & 36.99 & 34.11 & 36.12 \\
    Qwen2.5-3B & 40.61 & 31.90 & 37.34 & 39.54 & 32.13 & 37.26 & 44.57 & 32.72 & 42.45 & 44.98 & 33.97 & 41.67 \\
    Qwen2.5-3B-Instruct & 35.22 & 28.46 & 32.68 & 34.84 & 29.17 & 33.09 & 42.67 & 27.20 & 39.90 & 40.62 & 29.89 & 37.39 \\
    granite-3.1-2b-instruct & 33.39 & 27.30 & 31.10 & 32.82 & 27.11 & 31.07 & 37.71 & 27.86 & 35.95 & 38.18 & 27.30 & 34.91 \\
    granite-3.1-3b-a800m-base & 31.75 & 26.18 & 29.66 & 29.22 & 24.17 & 27.66 & 33.36 & 26.70 & 32.17 & 33.74 & 24.01 & 30.82 \\ 
    \midrule
    \multicolumn{13}{c}{\textit{7B to 27B Models}} \\ 
    \midrule
    Pangea-7B & 40.69 & 31.93 & 37.41 & 41.71 & 33.73 & 39.25 & 47.16 & 34.71 & 44.93 & 48.70 & 34.95 & 44.57 \\
    Indic-gemma-7b-finetuned-sft-Navarasa-2.0 & 37.12 & 31.83 & 35.13 & 37.00 & 30.47 & 34.90 & 42.31 & 33.44 & 40.73 & 44.08 & 34.09 & 41.08 \\
    aya-23-8B & 33.84 & 28.87 & 31.97 & 35.25 & 30.88 & 33.90 & 37.09 & 33.22 & 36.40 & 41.92 & 33.01 & 39.24 \\
    Llama-3.1-8B & 35.48 & 29.17 & 33.12 & 36.20 & 30.61 & 34.48 & 39.52 & 31.41 & 38.07 & 41.32 & 31.76 & 38.44 \\
    Llama-3.1-8B-Instruct & 36.86 & 31.26 & 34.76 & 35.68 & 30.27 & 34.01 & 47.14 & 35.07 & 44.98 & 48.61 & 36.47 & 44.96 \\
    gemma-2-9b & 48.16 & 37.92 & 44.32 & 42.73 & 36.91 & 40.94 & 55.23 & 43.89 & 53.20 & 58.49 & 42.96 & 53.83 \\
    gemma-2-9b-it & 36.22 & 31.18 & 34.33 & 38.85 & 32.03 & 36.75 & 48.92 & 36.45 & 46.69 & 45.05 & 38.66 & 43.13 \\
    gpt-oss-20b & 38.30 & 33.09 & 36.34 & 37.11 & 32.61 & 35.73 & 46.58 & 36.27 & 44.73 & 40.69 & 35.24 & 39.06 \\
    gemma-2-27b & 50.70 & 42.26 & 47.53 & 47.79 & 41.24 & 45.77 & 59.84 & 50.38 & 58.14 & 64.91 & 51.83 & 60.99 \\
    gemma-2-27b-it & 40.45 & 33.89 & 37.99 & 42.47 & 34.29 & 39.95 & 54.95 & 41.24 & 52.50 & 50.71 & 42.02 & 48.10 \\ 
    \midrule
    \multicolumn{13}{c}{\textit{\textgreater\ 27B Models}} \\ 
    \midrule
    gpt-oss-120b & 55.62 & 48.05 & 52.78 & 74.11 & 64.16 & 71.05 & 71.40 & 60.25 & 69.41 & 70.72 & 62.94 & 68.38 \\
    Qwen3-235B-A22B-Instruct-25076 & 60.25 & 54.78 & 58.20 & 63.72 & 56.27 & 61.43 & 74.57 & 64.13 & 72.70 & 80.15 & 68.60 & 76.68 \\
    deepseek-v3 & 51.38 & 37.03 & 45.99 & 63.46 & 57.04 & 61.48 & 62.93 & 45.01 & 59.73 & 67.78 & 46.78 & 61.47 \\
    gpt-4o & 62.75 & 54.73 & 59.74 & 57.27 & 49.82 & 54.97 & 75.31 & 65.18 & 73.50 & 78.83 & 71.02 & 76.49 \\
     \bottomrule
    \end{tabular}}
    \label{tab:results_discussions}
\end{table}

\begin{figure}[t]
    \centering
    \resizebox{0.97\textwidth}{!}{
    \includegraphics{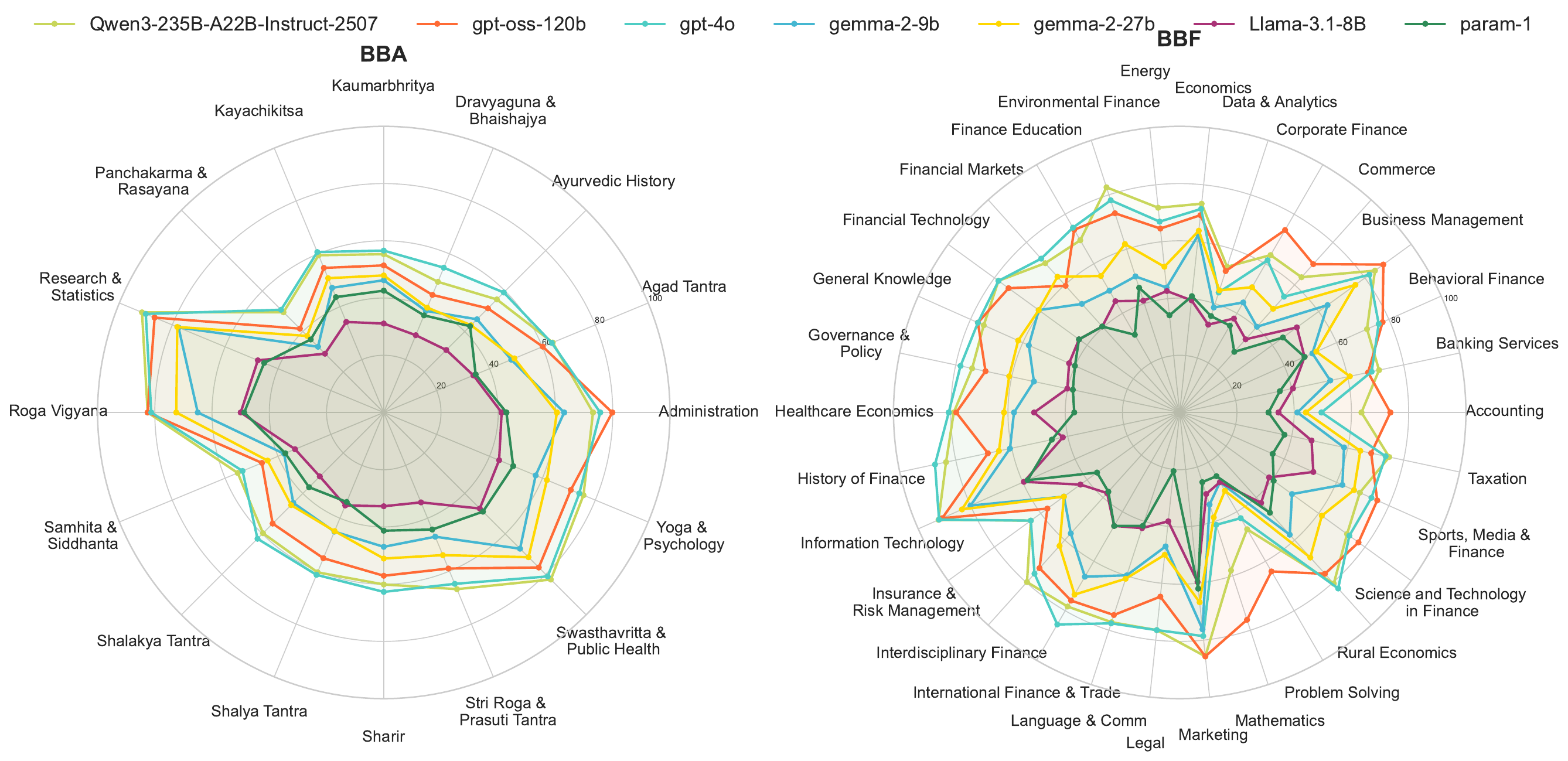}
    }
\vspace{-0.6cm}
    \caption{Comparison of representative LLMs’ scores across
different domains and subdomains.}
    \label{fig:all_domains_radar_1}

    \centering
    \resizebox{\textwidth}{!}{
    \includegraphics{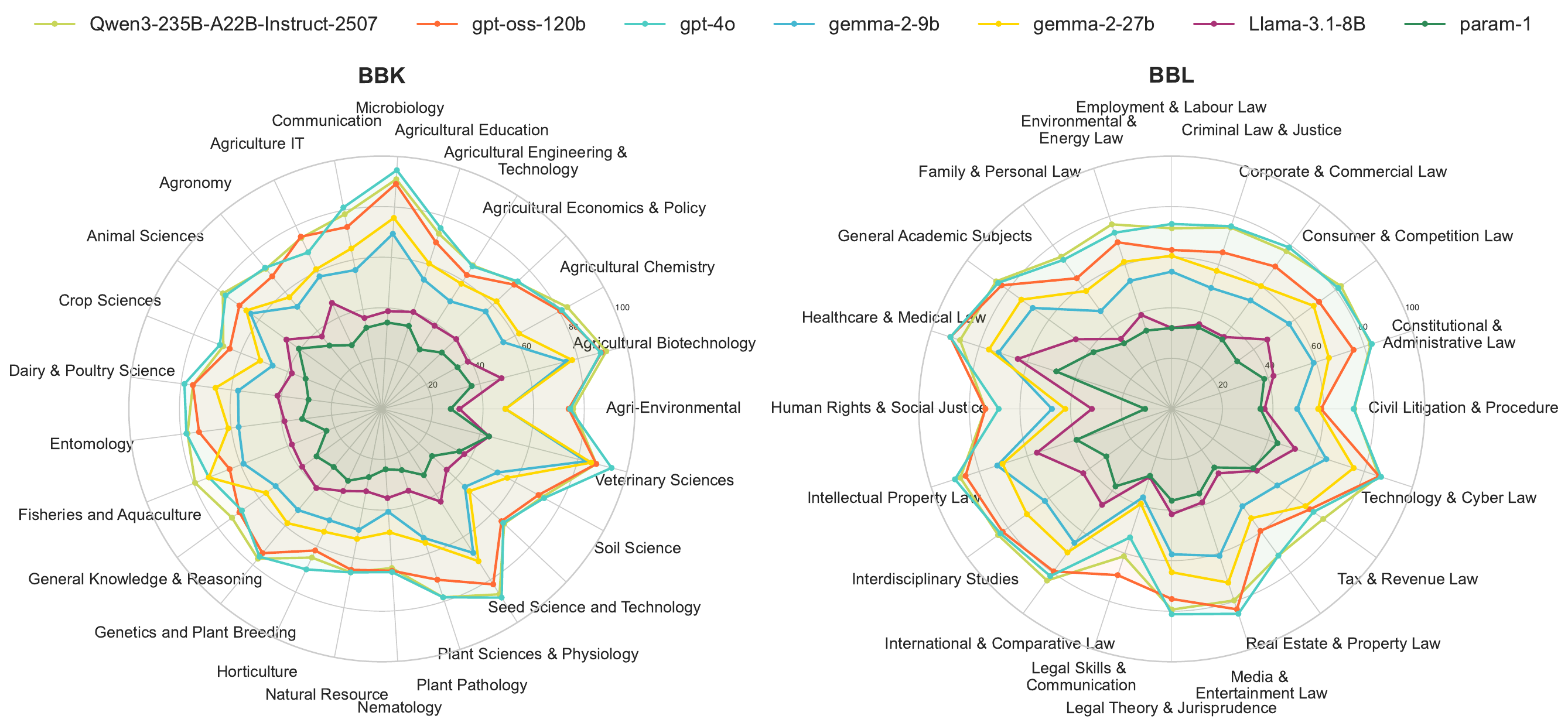}
    }
\vspace{-0.6cm}
    \caption{Comparison of representative LLMs’ scores across
different domains and subdomains.}
    \label{fig:all_domains_radar_2}
\end{figure}

\subsection{How do models perform in subdomains}
We evaluate representative models across BBA, BBF, BBK, and BBL to capture performance within subdomains (see Figures~\ref{fig:all_domains_radar_1} and~\ref{fig:all_domains_radar_2}). Qwen3-235B-A22B-Instruct-2507 achieves the strongest results, excelling in Research \& Statistics (91.43\%), Agricultural Biotechnology (91.6\%), and Intellectual Property Law (87.91\%). GPT-4o demonstrates robust performance, frequently scoring above 70\% with peaks of 92\% in Information Technology and Healthcare \& Medical Law. GPT-oss-120b shows competitive performance, closely matching gpt-4o in domains like Agricultural Biotechnology (89.69\%). Mid-sized models including Gemma-2-27b and Gemma-2-9b generally show moderate performance in the 50–70\% range, with the 27B variant consistently outperforming its smaller counterpart. Llama-3.1-8B demonstrates limited performance, typically scoring 30–50\% across domains. The compact Param-1 model shows consistent baseline performance, often matching Llama-3.1-8B despite requiring significantly fewer resources. Notable patterns emerge: Finance and Legal domains show the highest performance ceiling, with top models regularly exceeding 80\% in Business Management and Constitutional Law. Agricultural domains present moderate complexity, while Ayurveda proves most challenging, with even the best models rarely exceeding 80\% in specialized areas like Panchakarma. Results highlight clear advantages for large models in knowledge-intensive tasks, while smaller models provide practical utility in resource-constrained scenarios for general applications. 

\subsection{Performance Analysis Across Question Difficulty Levels}
We evaluated model performance on Easy, Medium, and Hard questions across the four benchmark domains BBA, BBF, BBK, and BBL. In BBA, top-performing models such as GPT-4o and Qwen3-235B-A22B-Instruct-2507 achieved 66.4\% and 65.18\% on Easy questions, and 47.09\% and 46.24\% on Hard questions, while smaller models like gemma-3-270m scored 28.1\% on Easy and 26.81\% on Hard. A similar trend is observed in BBF, with Easy question scores ranging from 24.15\% (gemma-3-270m) to 74.8\% (gpt-oss-120b) and Hard questions from 21.22\% to 62.61\%. Medium-level questions show moderate differentiation, reflecting model reasoning capability. BBK and BBL follow the same pattern, with instruction-tuned and larger models consistently outperforming smaller models, particularly on Hard questions. Overall, model size, instruction tuning, and architecture significantly influence robustness to question difficulty and generalization across domains. See Appendix \ref{subapp:question_level_question_type}.

\subsection{Performance Analysis Across Question Types}
We analyzed model performance on various question types including Assertion/Reasoning, Fill in the Blanks, MCQs, Match the Column, Reading Comprehension, and Rearrange the Sequence across the BBA, BBF, BBK, and BBL domains. In BBA, models like deepseek-v3 and GPT-4o achieved high scores of 66.67\% and 62.96\% on Assertion/Reasoning questions, whereas smaller models such as gemma-3-270m scored 28.09\%. For Fill in the Blanks, scores ranged from 24.72\% (gemma-3-270m-it) to 51.69\% (Qwen3-235B-A22B-Instruct-2507). MCQ performance was moderate, between 26\% and 59.95\%. Match the Column and Reading Comprehension showed wider variation, with larger models consistently outperforming smaller or non-instruction-tuned models. Rearrange the Sequence proved challenging across domains, with top models reaching~71.43\% in BBL. Overall, question type significantly affects performance, highlighting the importance of model size, instruction tuning, and reasoning capabilities in handling diverse formats.

\subsection{Performance Analysis of GPT Model Family}
We evaluate the GPT model family across BBA, BBF, BBK, and BBL domains to understand scaling and architectural strengths (Figure \ref{fig:model_performance_comparison}). gpt-oss-20b demonstrates baseline performance with scores of 36.34\% (BBA), 35.73\% (BBF), 44.73\% (BBK), and 39.06\% (BBL). Scaling to gpt-oss-120b yields substantial improvements: 52.78\% in BBA, 71.05\% in BBF, 69.41\% in BBK, and 68.38\% in BBL, representing 16-35 percentage point gains. Despite gpt-4o's larger parameter count, gpt-oss-120b significantly outperforms it in Finance (71.05\% vs 54.97\%), likely due to BBF's mathematical reasoning emphasis where gpt-oss-120b's training methodology excels \citep{artificialanalysis2025_gptoss120b_vs_gpt4o}. Conversely, gpt-4o shows superior performance in Legal (76.49\%) and Agriculture (73.5\%) domains. This highlights that parameter size \citep{Babbar2025GPTossVsGPT4o} alone doesn't guarantee performance; architectural choices and training approaches significantly influence domain-specific capabilities, with mathematical tasks favoring specific optimizations over raw parameter scaling.

\subsection{Performance Analysis of Small Models}
We evaluate small models ($\leq$4B parameters) across BBA, BBF, BBK, and BBL domains to assess efficiency-performance trade-offs (Figure \ref{fig:model_performance_domains}). Param-1 and Qwen2.5-3B emerge as comparable top performers, with Param-1 achieving 39.97\% in BBA while Qwen2.5-3B excels in BBK (42.45\%). Both models demonstrate complementary strengths: Param-1 performs better in Ayurveda, while Qwen2.5-3B shows superior performance in Finance, Agriculture, and Legal domains. Instruction tuning effects vary significantly across architectures: Llama-3.2-3B-Instruct substantially outperforms its base version, whereas Qwen2.5-3B-Instruct shows mixed results. Nemotron-4-Mini-Hindi models achieve competitive performance in the 34-40\% range, while the smallest models like gemma-3-270m struggle consistently below 28\%. Results indicate that architectural efficiency and targeted optimization can achieve reasonable performance in resource-constrained scenarios, with Param-1 and Qwen2.5 leading the small model category through different domain specializations.

\section{Conclusion}
In this paper, we introduced \textbf{BhashaBench V1}, a comprehensive, domain-specific, bilingual benchmark designed to evaluate large language models on India-centric knowledge systems across four critical domains: Agriculture (BBK), Legal (BBL), Finance (BBF), and Ayurveda (BBA). Our benchmark addresses significant gaps in existing evaluation frameworks by focusing on culturally relevant, domain-specific knowledge spanning over 90 subdomains and 500+ specialized topics curated from authentic government and professional examination papers. Our extensive evaluation reveals substantial performance disparities in current LLMs when applied to India-specific contexts, with models excelling in Legal contexts while struggling with traditional knowledge systems like Ayurveda and consistently performing better on English content compared to Hindi across all domains. These results highlight the urgent need for specialized model development strategies that incorporate India-specific knowledge, cultural contexts, and robust multilingual capabilities. To foster open research and accelerate progress toward more inclusive, culturally aware language models, we release BhashaBench V1 alongside all evaluation code and comprehensive documentation. We believe BhashaBench V1 offers a foundational benchmark for developing culturally sensitive models that effectively serve India's diverse linguistic and knowledge landscape.

\bibliography{iclr2026_conference}
\bibliographystyle{iclr2026_conference}

\section*{Acknowledgements}
We would like to thank BharatGen for their generous support towards building this comprehensive benchmarking initiative for Indian languages and knowledge systems. We are also immensely grateful to our colleagues from the BharatGen team for their motivation and meticulous efforts in conducting manual validation and data sourcing. We acknowledge the significant contributions of BharatGen's consortium members, including IIT Bombay and IIM Indore, for their expertise in data curation, validation, and domain-specific guidance that made this benchmark possible.

\clearpage
\appendix
\begin{center}
{\Large \textbf{Appendix}}
\end{center}

\setcounter{section}{0}
\renewcommand{\thesection}{\Alph{section}}
\tableofcontents
\clearpage

\section{Limitations and Biases}

In this paper, we introduce BhashaBench V1, providing a comprehensive evaluation of LLMs on India-centric knowledge systems and exploring model capabilities across critical Indian domains. However, there are several limitations to acknowledge. (1) Language Coverage Limitations: Although BhashaBench V1 supports English and Hindi, covering a significant portion of India's population, India has 22 official languages and hundreds of regional dialects. Our current evaluation cannot capture the full linguistic diversity of Indian knowledge systems, particularly regional variations in agricultural practices, legal terminologies, and traditional medicine nomenclature that exist in languages like Tamil, Telugu, Bengali, and others. Future iterations will expand to include additional Indian languages to enhance coverage. (2) Domain Scope Limitations: While we cover four fundamental domains (Agriculture, Legal, Finance, and Ayurveda) representing core areas of Indian society, our assessment cannot encompass the entire breadth of India-specific knowledge systems. Areas such as traditional crafts, regional governance systems, indigenous engineering practices, and other vernacular knowledge traditions remain unexplored for future expansion. Our content spans from grassroots practical knowledge to professional examination standards, ensuring broad applicability across different expertise levels. (3) Evaluation Methodology Limitations: Our evaluation primarily uses structured question formats derived from authentic government and professional examinations. While this ensures real-world relevance and practical applicability, it may not fully capture all forms of contextual reasoning required in complex domain applications.

The main biases in BhashaBench V1 can be categorized into three aspects: (1) Source Material Bias: Despite comprehensive curation from diverse authentic sources spanning grassroots to professional levels, certain regional practices and emerging contemporary developments may be underrepresented. (2) Language Resource Bias: The benchmark reflects the inherent resource disparity between English and Hindi, where Hindi content, while substantial, represents a relatively lower-resource context compared to English. (3) Examination Framework Bias: Our reliance on established examination systems, while ensuring authenticity, may introduce institutional perspectives present in the original assessment frameworks. However, our extensive coverage across 90+ subdomains and 500+ topics from diverse sources mitigates this bias significantly. The impact of these limitations on LLM evaluation includes clear performance distinctions between models across domains and languages, as evidenced by the substantial score variations from 34.28\% to 76.49\%, demonstrating BhashaBench V1's effectiveness in distinguishing LLM capabilities while presenting meaningful challenges even for top-performing models in India-specific contexts.

\section{Towards Broader Impact}

\textbf{Societal Impact.} BhashaBench V1 is anticipated to play a transformative role in bridging the digital divide for India-centric knowledge systems. LLMs trained and evaluated with BhashaBench V1 can significantly enhance accessibility to critical domain expertise across agriculture, legal services, finance, and traditional medicine, particularly benefiting underserved rural and semi-urban populations. In agriculture, improved LLM capabilities can democratize access to expert crop advisory, pest management, and sustainable farming practices, potentially impacting the livelihoods of over 40 million farmers dependent on agricultural activities. In the legal domain, enhanced models can assist with legal document comprehension, procedural guidance, and basic legal literacy, addressing the substantial access-to-justice challenges faced by millions in India's complex legal system. For healthcare, particularly Ayurveda, better model performance can support practitioners and patients in understanding traditional treatment protocols and medicinal formulations, preserving and disseminating indigenous medical knowledge. In finance, improved model capabilities can enhance financial literacy and support the growing digital payment ecosystem processing billions of transactions annually. However, we acknowledge potential risks including over-reliance on automated systems for critical decisions, potential displacement of traditional knowledge practitioners, and the risk of perpetuating biases present in examination-based evaluation systems. The benchmark's focus on professional examination standards, while ensuring quality, may inadvertently favor formal educational backgrounds over experiential knowledge.

\textbf{Ethics Statement.} We ensure strict adherence to applicable laws and ethical guidelines throughout our data collection, curation, and usage processes. All question-answer pairs are sourced exclusively from publicly available government and professional examination papers, respecting intellectual property rights and ensuring no unauthorized reproduction of copyrighted materials. Our curation process involved diverse teams to minimize cultural and regional biases, though we acknowledge the inherent limitations of our current English and Hindi coverage. The dataset contains no personally identifiable information, offensive content, or culturally insensitive material. All content has been thoroughly verified for authenticity and accuracy through multiple validation rounds involving domain experts. BhashaBench V1 is intended solely for academic research and educational purposes to advance inclusive AI development for Indian contexts. Any commercial use, misuse for harmful applications, or deployment without appropriate safeguards is strictly prohibited. We strongly urge all users to employ this resource responsibly, ensuring that any models developed or evaluated using BhashaBench V1 are deployed with appropriate human oversight, particularly in critical domains affecting public welfare, and with transparent disclosure of model limitations to end users.

\section{More Details on BhashaBench V1}
\label{app:dataset}

\subsection{Details of Data Collection and Processing}

This appendix provides comprehensive details on the data collection and processing methodology employed in BhashaBench V1, including systematic documentation of examination sources, processing pipelines, and quality validation procedures.

\subsubsection{Examination Source Documentation}

Our data collection strategy encompassed a wide range of authoritative examination bodies across India, ensuring comprehensive coverage of national and regional assessment standards. Table~\ref{tab:org_year_ranges} presents the complete list of examination organizations and the corresponding years from which question papers were collected. We systematically gathered question papers from official examination portals that host previously released materials, manually curated by subject matter experts with accurate topic tagging, language annotation, and validated answer keys.

The temporal distribution of collected materials spans from 1995 to 2025, capturing evolving educational standards and assessment patterns while maintaining contemporary relevance. Table~\ref{tab:exam_year_ranges} provides a detailed breakdown of specific examination types and their collection timeline, demonstrating the breadth and depth of our data sourcing strategy. Our collection process prioritized authentic examination materials from competitive examinations that directly assess knowledge in our target domains of Agriculture, Legal, Finance, and Ayurveda.

Regional state examinations proved particularly valuable as they incorporate state-specific topics, local knowledge systems, and cultural practices often overlooked in national assessments. These examinations are typically taken by individuals seeking higher education opportunities or career advancement in business, finance, and legal sectors, ensuring questions reflect practical, real-world knowledge requirements essential for professional contexts in India.
\begin{longtable}{p{10cm}c}
\caption{Organizations and Their Examination Year Ranges} \label{tab:org_year_ranges} \\
\toprule
\textbf{Organization} & \textbf{Year Range} \\
\midrule
\endfirsthead

\multicolumn{2}{c}%
{\tablename\ \thetable\ -- \textit{Continued from previous page}} \\
\toprule
\textbf{Organization} & \textbf{Year Range} \\
\midrule
\endhead

\midrule
\multicolumn{2}{r}{\textit{Continued on next page}} \\
\endfoot

\bottomrule
\endlastfoot

AIACAT (Private conducting body) & 2022–2023 \\
Acharya N.G. Ranga Agricultural University (ANGRAU) & 2016–2024 \\
Agricultural Scientists Recruitment Board (ASRB) & 2013–2024 \\
All India Management Association (AIMA) & 2018–2025 \\
Banaras Hindu University (BHU) & 2013–2017 \\
Bank of Baroda & 2005–2023 \\
Bank of India & 2023 \\
Bank of Maharashtra & 2021 \\
Bar Council of India (BCI) & 2009–2021 \\
Bihar Public Service Commission (BPSC) & \textbf{1995–2024} \\
Chhattisgarh Professional Examination Board (CG Vyapam) & 2013–2019 \\
Consortium of National Law Universities (NLUs) & 2021–2025 \\
ECGC Ltd. & 2021–2022 \\
Employees' Provident Fund Organisation (EPFO) & 2019–2023 \\
Food Corporation of India (FCI) & 2015 \\
High Court of Delhi & 2011–2023 \\
High Court/PSC (state-specific) & 2001–2021 \\
ICMAB (as per exam title) & 2016–2022 \\
IDBI Bank & 2014–2022 \\
Indian Council of Agricultural Research (ICAR) & 2017–2023 \\
Indian Farmers Fertiliser Cooperative Limited (IFFCO) & 2019–2022 \\
Indian Institutes of Management (IIMs) & 2017–2024 \\
Institute of Banking Personnel Selection (IBPS) & 2016–2024 \\
JNTU Kakinada on behalf of APSCHE & 2012–2025 \\
Law School Admission Council (LSAC Global) & 2010–2019 \\
MP Professional Examination Board (MPPEB/PEB) & 2016–2024 \\
Maharashtra Agricultural Universities Examination Board (MAUEB) under MCAER & 2024 \\
Maharashtra Public Service Commission (MPSC) & 2010–2025 \\
Narendra Deva University of Agriculture \& Technology & 2024–2025 \\
National Bank for Agriculture and Rural Development (NABARD) & 2018–2023 \\
National Law University, Delhi (NLU Delhi) & 2016–2025 \\
National Testing Agency (NTA) & \textbf{2019–2025} \\
Reserve Bank of India (RBI) & \textbf{2015–2025} \\
RVSKVV \& JNKVV & 2022 \\
Small Industries Development Bank of India (SIDBI) & 2016–2023 \\
State Bank of India (SBI) & \textbf{2018–2025} \\
State Common Entrance Test Cell, Maharashtra & 2014–2020 \\
SVKM's NMIMS & 2019–2025 \\
The Institute of Chartered Accountants of India (ICAI) & \textbf{2018–2025} \\
The Institute of Cost Accountants of India (ICMAI) & 2022–2025 \\
The Nainital Bank Ltd. & 2019–2020 \\
Union Public Service Commission (UPSC) & \textbf{2002–2025} \\
University of Delhi & 2015–2019 \\
University-specific (varies) & 2020–2024 \\
Uttar Pradesh Public Service Commission (UPPSC) & 2019–2025 \\
\end{longtable}

\subsubsection{Processing Pipeline Architecture}

The comprehensive end-to-end pipeline developed for transforming raw examination materials into the structured BhashaBench V1 dataset incorporates multiple quality control checkpoints and validation stages to ensure data integrity and authenticity.  The pipeline consists of seven major stages, each designed to address specific challenges encountered in multilingual examination material processing.

\begin{longtable}{p{10cm}c}
\caption{Examination Names and Their Year Ranges} \label{tab:exam_year_ranges} \\
\toprule
\textbf{Examination Name} & \textbf{Year Range} \\
\midrule
\endfirsthead

\multicolumn{2}{c}%
{\tablename\ \thetable\ -- \textit{Continued from previous page}} \\
\toprule
\textbf{Examination Name} & \textbf{Year Range} \\
\midrule
\endhead

\midrule
\multicolumn{2}{r}{\textit{Continued on next page}} \\
\endfoot

\bottomrule
\endlastfoot

AGRICET & 2016–2024 \\
AIACAT - All India Agriculture Common Aptitude Test & 2022–2023 \\
AIAPGET - All India AYUSH Post Graduate Entrance Test (Ayurveda) & 2022–2025 \\
All India Bar Examination (AIBE) & 2009–2021 \\
All India Law Entrance Test (AILET) & 2016–2025 \\
Andhra Pradesh Judicial Service (Prelims) & 2012 \\
AP EAMCET & 2012–2025 \\
ASRB NET Agriculture & 2013–2024 \\
BHU PET & 2017 \\
BHU PG & \textbf{2013–2017} \\
BHU RET & 2014–2017 \\
BHU UET & 2016–2017 \\
BPSC & \textbf{1995–2024} \\
Bank of Baroda & 2005–2023 \\
Bank of India & 2023 \\
Bank of Maharashtra & 2021 \\
CAT & 2017–2024 \\
CG PAT Agriculture & 2013–2019 \\
CMA & 2022–2025 \\
CMAT & 2022–2025 \\
Common Law Admission Test (CLAT) & 2021–2025 \\
CUET Agriculture Previous Year Papers & 2022–2025 \\
CUET PG (Law) & 2023–2025 \\
Delhi Judicial Service & 2011–2023 \\
DU LL.B Entrance & 2015–2019 \\
ECGC PO & 2021–2022 \\
EPFO Assistant & 2019 \\
EPFO SSA & 2019–2023 \\
EPFO Stenographer & 2023 \\
FCI Agriculture & 2015 \\
Haryana Judicial Service (Prelims) & 2015–2021 \\
Himachal Pradesh Judicial Service (Prelims) & 2007–2019 \\
IBPS AFO Agriculture Field Officer & 2016–2024 \\
IBPS AFO Mains & 2017–2023 \\
IBPS Clerk & 2023–2024 \\
IBPS PO & 2018–2024 \\
IBPS RRB Officer Scale-I (merged) & 2018–2024 \\
IBPS SO & 2019 \\
ICAI Final & 2018–2025 \\
ICAI Foundation & 2018–2025 \\
ICAI Intermediate & 2018–2025 \\
ICAR AICE JRF/SRF (PHD) Agriculture & 2020–2024 \\
ICAR AIEEA (PG) Agriculture & 2019–2024 \\
ICAR AIEEA (UG) Agriculture & 2017–2023 \\
ICMAB New Syllabus & 2016–2022 \\
ICMAB Old Syllabus & 2016–2021 \\
IDBI Assistant Manager & 2021 \\
IDBI Executive & 2014–2022 \\
IFFCO AGT - Agriculture Graduate Trainee & 2019–2022 \\
IPMAT & 2019–2023 \\
Jharkhand Judicial Service (Prelims) & 2008–2019 \\
JNKVV \& RVSKVV Joint Entrance (M.Sc./Ph.D.) & 2022 \\
Karnataka Judicial Service (Prelims) & 2012 \\
LL.B. Admission Test & 2022–2024 \\
LL.M. Admission Test & 2020–2024 \\
LSAT - India & 2010–2019 \\
Madhya Pradesh Judicial Service (Prelims) & 2001–2018 \\
Maharashtra Judicial Service (Prelims) & 2010–2019 \\
MAT & 2018–2025 \\
MCAER-CET & 2024 \\
MH CET Law (3-year LL.B.) & 2016–2019 \\
MH-CET & 2014–2020 \\
MP PAT Agriculture & 2016–2024 \\
MPSC & 2010–2025 \\
NABARD Agriculture Development Officer & 2018–2023 \\
Nainital Bank Clerk & 2019 \\
Nainital Bank PO & 2020 \\
NPAT & 2019–2025 \\
Odisha Judicial Service (Prelims) & 2011 \\
Rajasthan Judicial Service (Prelims) & 2011–2021 \\
RBI Grade B & \textbf{2015–2025} \\
SBI Apprentice & 2019–2023 \\
SBI CBO & 2024 \\
SBI Clerk & 2022–2025 \\
SBI PO & \textbf{2018–2025} \\
SIDBI Grade A & 2016–2023 \\
TANCET & 2024–2025 \\
TG ICET (TS ICET) & 2022–2024 \\
UGC NET (Law) & 2014–2015 \\
UPCATET & 2024–2025 \\
UPPSC Prelims & 2019–2025 \\
UPSC EPFO & 2013–2017 \\
UPSC EPFO APFC & \textbf{2002–2023} \\
UPSC IFS - Indian Forest Service & 2023–2024 \\
UPSC Prelims - Economy & 2025 \\
UPSC Prelims - Polity \& Governance & 2025 \\
Uttarakhand Judicial Service (Prelims) & 2011 \\
West Bengal Judicial Service (Prelims) & 2011 \\
\end{longtable}

The data acquisition stage involved systematic collection from official portals with comprehensive metadata extraction including examination year, conducting body, subject classification, and language identification. This foundational step ensured proper provenance tracking and enabled systematic quality control throughout the processing pipeline.

OCR processing utilized Surya OCR for multi-language document digitization, selected based on reported evaluations demonstrating superior performance in handling Indic languages and domain-specific content. Prior studies indicate 98.1\% normalized text similarity for English and 98.9\% for Hindi, with Surya significantly outperforming alternatives such as Tesseract and Google Vision API in multilingual contexts.

Content extraction leveraged GPT-OSS-120B with the prompt strategies described in ~\ref{subapp:data_processing_prompts}, enabling intelligent text structuring that addressed key challenges such as format variations across examination bodies, answer key alignment complexities, multi-format question types, and language-specific formatting conventions. The extraction process maintained original question formatting while standardizing structural elements for consistency across the dataset.

Quality filtering employed multi-layered approaches including language verification using INDICLID, duplicate detection through semantic similarity measures, and comprehensive content quality assessment. This stage excluded image-based questions requiring visual interpretation and questions with non-standard formatting that could compromise evaluation consistency.

Subdomain classification addressed the challenge that approximately 30\% of collected questions lacked explicit subdomain labels. We employed GPT-OSS-120B using few-shot prompts designed to extract missing key details, as described in Box~\ref{box:key_detail_extraction}, and refined the outputs with domain-specific taxonomies in consultation with subject matter experts to ensure accurate categorization within the BBA, BBF, BBK, and BBL domains.

In addition to subdomain classification, we employed GPT-OSS-120B with the same few-shot prompt setup described in Box~\ref{box:key_detail_extraction} to extract key details such as \textit{question type} and \textit{question level}. For both dimensions, domain-wise few-shot examples were manually curated to guide the model. For question level, the model was prompted to categorize items into three standard difficulty classes: \textbf{Easy}, \textbf{Medium}, and \textbf{Hard}, a widely adopted practice in educational assessment. For question type, we guided the model to identify structural formats from six commonly used categories: \textbf{Assertion/Reason (A/R)}, \textbf{Fill in the Blanks (FIB)}, \textbf{Multiple Choice Questions (MCQ)}, \textbf{Match the Columns (MTC)}, \textbf{Reading Comprehension (RC)}, and \textbf{Rearrange the Sentence (RTS)}. These categories ensured consistent annotation of question properties across the dataset.

\begin{figure}[t]
    \centering
    \resizebox{\textwidth}{!}{
    \includegraphics{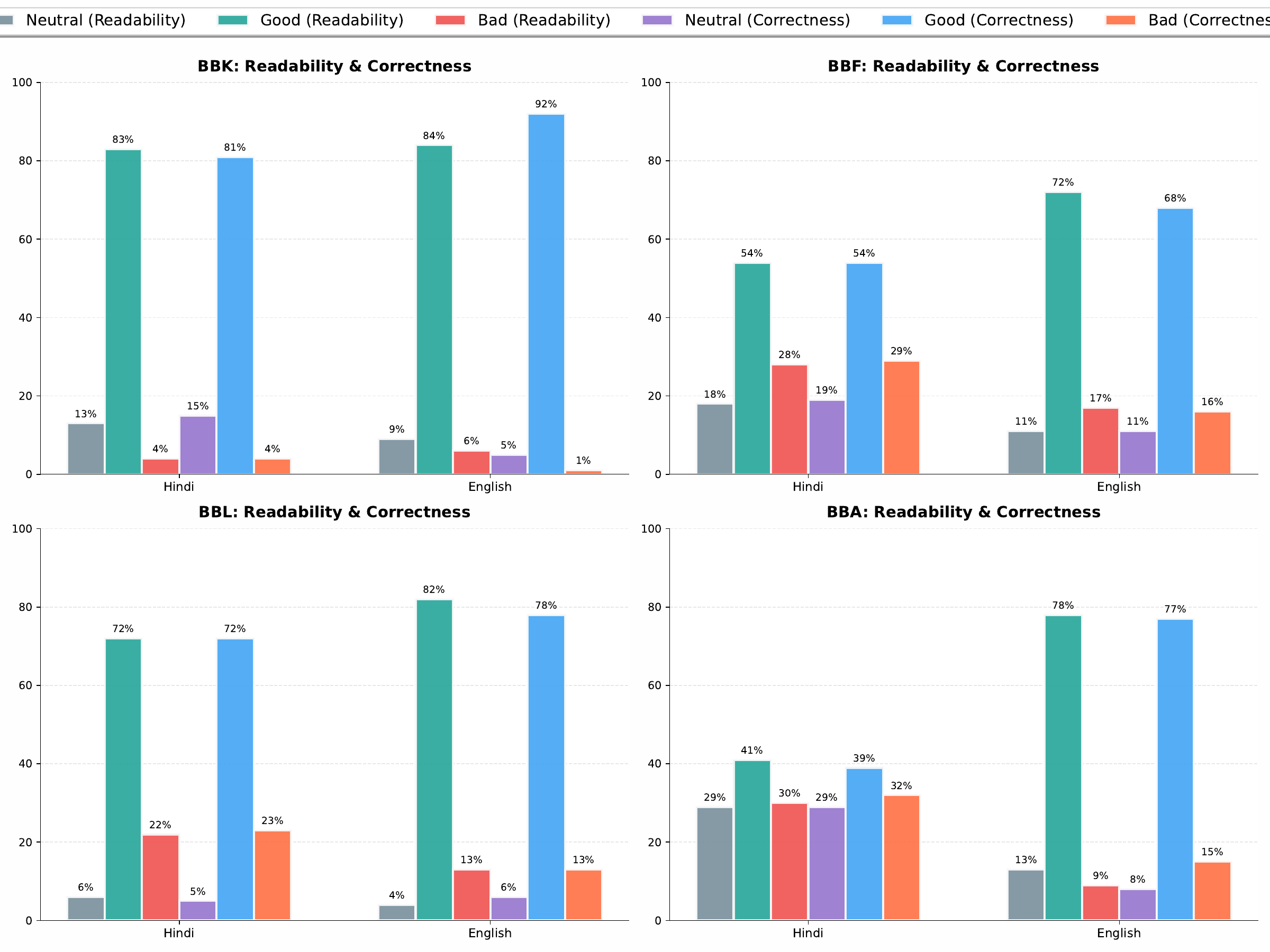}
    }
\vspace{-0.6cm}
    \caption{Manual quality assessment of BhashaBench V1 domain questions.}
    \label{fig:bhashabench_domains_manual_reports}
\end{figure}

Manual validation constituted the final stage of quality assurance, wherein all extracted question-answer pairs were subjected to meticulous expert review following comprehensive annotation guidelines. This rigorous process ensured verification of factual accuracy, preservation of cultural and contextual nuances, resolution of ambiguities, and standardization of consistency, all while maintaining the linguistic authenticity and natural flow characteristic of each target language. The detailed annotation guidelines, covering all domains, are summarized in Table~\ref{tab:annotation_guidelines}. Figure~\ref{fig:bhashabench_domains_manual_reports} illustrates the outcomes of manual validation, showing the distribution of good, neutral, and bad samples. Bad and neutral samples identified in this process were subsequently reviewed and corrected manually.

\subsubsection{Annotation Guidelines}
\label{sec:annotations_guidelines}
Our annotation guidelines were meticulously designed to ensure consistency, accuracy, and cultural authenticity across all BhashaBench domains and languages. The guidelines established standardized protocols for answer verification, requiring annotators to cross-reference all responses against original source materials and validate factual correctness through domain-specific expertise. Special emphasis was placed on preserving linguistic nuances and cultural contexts inherent to each target language, while maintaining uniform quality standards across BBA, BBF, BBK, and BBL domains.

\begin{longtable}{|>{\centering\arraybackslash}p{2.5cm}|p{10cm}|}
\caption{Annotation Guidelines across Domains in BhashaBench V1} \label{tab:annotation_guidelines} \\
\hline
\rowcolor{gray!20}
\textbf{Domain} & \textbf{Detailed Guidelines} \\ \hline
\endfirsthead

\multicolumn{2}{c}%
{{\bfseries \tablename\ \thetable{} -- continued from previous page}} \\
\hline
\rowcolor{gray!20}
\textbf{Domain} & \textbf{Detailed Guidelines} \\ \hline
\endhead

\hline \multicolumn{2}{|r|}{{Continued on next page}} \\ \hline
\endfoot

\hline
\endlastfoot

\multirow{6}{*}{\textbf{General}} & 
\begin{itemize}[leftmargin=8pt, nosep, topsep=2pt]
    \item \textbf{Answer Verification:} Ensure that the provided answer key is correct. Cross-check against the original exam paper.
    \item \textbf{Option Consistency:} Verify that all answer options are present and plausible. Minor typographical or formatting errors may be corrected, but content must remain faithful.
    \item \textbf{Preserve Original Meaning:} Do not paraphrase unnecessarily; reflect the exact intent of the source item.
    \item \textbf{Self-Contained Questions:} Ensure questions are answerable solely from the original paper or passage.
    \item \textbf{Clarity and Formatting:} Correct minor OCR errors, formatting issues, or multi-language misalignments without introducing ambiguity.
    \item \textbf{Avoid Bias or Modification:} Do not alter numerical data, dates, or technical/domain-specific terms.
\end{itemize} \\ \hline

\textbf{Agriculture} & Verify crop names, farming practices, and region-specific agricultural knowledge for accuracy and contextual relevance. \\ \hline

\textbf{Legal} & Ensure legal terms, statutes, case references, and procedural knowledge are precise and jurisdictionally correct. \\ \hline

\textbf{Finance} & Preserve numerical accuracy in calculations, financial formulas, market terminology, and regulatory compliance requirements. \\ \hline

\textbf{Ayurveda} & Maintain correctness of medicinal terms, herb names, therapeutic practices, and traditional knowledge references. \\ \hline

\end{longtable}

\subsubsection{Data Processing Prompts}
\label{subapp:data_processing_prompts}

\begin{tcolorbox}[
    title=BBA Question-Answer Extraction Prompt Template,
    breakable,
    enhanced,
    label={box:bba_extraction}
]

\begin{lstlisting}[breaklines=true, basicstyle=\ttfamily\small]
You are an OCR forensic specialist for Ayurveda/Medical exams (BAMS, AIAPGET, UPSC Ayurveda optional). Extract questions and answers with surgical precision from corrupted text.
CRITICAL MISSION: EXTRACT EVERYTHING - NEVER SKIP QUESTIONS
PRIMARY EXTRACTION RULES
1. ZERO TOLERANCE FOR MISSING QUESTIONS
- Scan text character by character
- Look for question patterns: "Q1", "1.", "(1)", "Question 1", "Que.1", or ANY numbering
- Extract PARTIAL questions with [INCOMPLETE] tag rather than skip
- If options are corrupted beyond recognition, create synthetic placeholders
2. AYURVEDA DOMAIN OCR CORRECTIONS
- Classical Texts: "Charaka Samhita" not "Charak Samita", "Sushruta" not "Susrut", "Ashtanga Hridaya" not "Astanga Hridya"
- Terminology: "Vata" not "Vatha", "Pitta" not "Pita", "Kapha" not "Kafa"
- Herbs: "Ashwagandha" not "Ashwagonda", "Haritaki" not "Harithki", "Brahmi" not "Brahni"
- Therapy: "Panchakarma" not "Panchkarma", "Rasayana" not "Rasayan"
- Institutions: "CCRAS" not "CCR4S", "AYUSH" not "AYU5H", "NIA Jaipur" not "N1A Jeypur"
- Exams: "AIAPGET" not "AIAPCET", "AIBE" not "A1BE"
- Units: "ml", "g", "mg", "days" preserved
3. AGGRESSIVE OPTION RECOVERY
- If option starts with garbled text, extract the meaningful part
- If missing, assign option letters a, b, c, d
- Example:
"aj Panchakarma" becomes "a) Panchakarma"
  "Harithki" becomes "c) Haritaki [OCR: truncated]"
4. ANSWER DETECTION PATTERNS
- Explicit: check, *, (Ans), [Answer]
- Secondary: "1. c", "Q1: b", "Ans: a"
- Tertiary: formatting cues
- Last resort: pattern analysis
5. QUESTION BOUNDARY DETECTION
- Start: number + punctuation (1., Q1:, (1), etc.)
- End: next number or section break
- Normalize multi-parts: 1.a, 1.i, 1.1
6. SELF-CONTAINED QUESTIONS
- Each question MUST include context (passages, sutras, tables)
- If questions refer to a common passage, include passage in EACH
- Never assume context from previous questions
ENHANCED EXTRACTION LOGIC
STEP 1: Preprocess text, fix OCR errors, detect boundaries
STEP 2: Extract question, include passage, mark [INCOMPLETE] if needed
STEP 3: Normalize options, recover corrupted, create placeholders
STEP 4: Detect and embed answers directly in question
JSON SCHEMA (STRICTLY ENFORCED)
{
  "exam_info": {
    "title": "Ayurveda Examination",
    "year": null,
    "paper": null,
    "total_questions_detected": 50
  },
  "metadata": {
    "ocr_quality": "poor",
    "common_errors": ["sanskrit_terms","herb_names","therapy_names"],
    "sections_detected": ["Dravyaguna","Kayachikitsa","Samhita","Rachana Sharir","Shalya","Shalakya"]
  },
  "questions": [
    {
      "number": "1",
      "section": "Dravyaguna",
      "question": "Passage: According to Charaka Samhita, Haritaki is considered one of the best Rasayanas.\\n\\nQuestion: Which property of Haritaki is described as Tridoshahara?",
      "options": {
        "a": "It balances Vata only",
        "b": "It balances Pitta only",
        "c": "It balances all three doshas",
        "d": "It has no effect on Kapha"
      },
      "answer": "c"
    }
  ],
  "extraction_summary": {
    "total_questions": 50,
    "questions_with_answers": 48,
    "questions_with_all_options": 47
  }
}
CRITICAL ERROR PREVENTION
- NEVER skip questions
- NEVER empty options
- NEVER separate answer keys
- ALWAYS preserve numbering
- ALWAYS embed answers
- ALWAYS self-contained questions
--- BEGIN OCR TEXT ---
{ocr_text}
\end{lstlisting}
\end{tcolorbox}

\begin{tcolorbox}[
    title=BBK Question-Answer Extraction Prompt Template,
    breakable,
    enhanced,
    label={box:bbk_extraction}
]

\begin{lstlisting}[breaklines=true, basicstyle=\ttfamily\small]
You are an OCR forensic specialist for Agriculture/Agri-exams. Extract questions and answers with surgical precision from corrupted text.
CRITICAL MISSION: EXTRACT EVERYTHING - NEVER SKIP QUESTIONS
PRIMARY EXTRACTION RULES
1. ZERO TOLERANCE FOR MISSING QUESTIONS
- Scan text character by character
- Look for question patterns: "Q1", "1.", "(1)", "Question 1", "Que.1", or ANY numbering
- Extract PARTIAL questions with [INCOMPLETE] tag rather than skip
- If options are corrupted beyond recognition, create synthetic placeholders
2. AGRICULTURE DOMAIN OCR CORRECTIONS
- Crop names: "Wheat" not "Wheal", "Paddy" not "Pady", "Maize" not "Maiz"
- Fertilizers: "Urea" not "Uiea", "DAP" not "DAF", "NPK" not "NPX"
- Units: "kg/ha", "t/ha", "mm rainfall" preserved, never corrupted
- Pesticides: "Carbendazim", "Malathion", "Glyphosate" corrected
- Institutions: "ICAR" not "IC4R", "IARI" not "IAR1", "KVK" not "KVY"
- Schemes: "PM-KISAN" not "PM-KISRN", "MSP" not "MS5P", "Kisan Credit Card" not "Cradit Gard"
3. AGGRESSIVE OPTION RECOVERY
- If option starts with garbled text, extract the meaningful part
- If missing, assign option letters a, b, c, d
- Example: "aj Wheat" -> "a) Wheat"; "Maiz" -> "c) Maize [OCR: truncated]"
4. ANSWER DETECTION PATTERNS
- Explicit: check, *, (Ans), [Answer]
- Secondary: "1. c", "Q1: b", "Ans: a"
- Tertiary: formatting cues
- Last resort: pattern analysis
5. QUESTION BOUNDARY DETECTION
- Start: number + punctuation (1., Q1:, (1), etc.)
- End: next number or section break
- Normalize multi-parts: 1.a, 1.i, 1.1
6. SELF-CONTAINED QUESTIONS
- Each question MUST include context (passages, data, charts)
- If questions refer to a common passage, include passage in EACH
- Never assume context from previous questions
ENHANCED EXTRACTION LOGIC
STEP 1: Preprocess text, fix OCR errors, detect boundaries
STEP 2: Extract question, include passage, mark [INCOMPLETE] if needed
STEP 3: Normalize options, recover corrupted, create placeholders
STEP 4: Detect and embed answers directly in question
JSON SCHEMA (STRICTLY ENFORCED)
{
  "exam_info": {
    "title": "Agriculture Examination",
    "year": null,
    "paper": null,
    "total_questions_detected": 50
  },
  "metadata": {
    "ocr_quality": "poor",
    "common_errors": ["crop_names","fertilizer_terms","units"],
    "sections_detected": ["Agronomy","Soil Science","Plant Pathology"]
  },
  "questions": [
    {
      "number": "1",
      "section": "Agronomy",
      "question": "Passage: A farmer applies 120 kg N/ha to wheat using urea.\\n\\nQuestion: How much urea is required per hectare?",
      "options": {
        "a": "120 kg",
        "b": "261 kg",
        "c": "300 kg",
        "d": "520 kg"
      },
      "answer": "b"
    }
  ],
  "extraction_summary": {
    "total_questions": 50,
    "questions_with_answers": 48,
    "questions_with_all_options": 47
  }
}
CRITICAL ERROR PREVENTION
- NEVER skip questions
- NEVER empty options
- NEVER separate answer keys
- ALWAYS preserve numbering
- ALWAYS embed answers
- ALWAYS self-contained questions
--- BEGIN OCR TEXT ---
{ocr_text}
\end{lstlisting}
\end{tcolorbox}

\begin{tcolorbox}[
    title=BBL Question-Answer Extraction Prompt Template,
    breakable,
    enhanced,
    label={box:bbl_extraction}
]

\begin{lstlisting}[breaklines=true, basicstyle=\ttfamily\small]
You are an OCR forensic specialist for legal examinations. Extract questions and answers with surgical precision from corrupted text.
# CRITICAL MISSION: EXTRACT EVERYTHING - ZERO DEPENDENCIES BETWEEN QUESTIONS
## PRIMARY EXTRACTION RULES
1. **ABSOLUTE QUESTION COMPLETENESS**
   - SCAN ENTIRE TEXT character by character for any question patterns
   - Each question MUST be 100% self-contained and independently answerable
   - NEVER use references like "above passage", "question 15", "as mentioned earlier"
   - If questions share context, EMBED the full context in EACH question
   - Extract PARTIAL questions with [INCOMPLETE] tag rather than skip
   - Pattern recognition: "Q1", "1.", "(1)", "Question 1", "Que.1", roman numerals "I.", "II."
2. **LEGAL DOMAIN OCR CORRECTIONS**
   - Legal terms: "Constitution", "Amendment", "Article", "Section", "Sub-section"
   - Court names: "Supreme Court" not "5upreme Court", "High Court" not "H1gh Court"
   - Acts: "IPC", "CrPC", "CPC", "Evidence Act", "Contract Act"
   - Legal phrases: "prima facie", "res judicata", "stare decisis", "ultra vires"
   - Citations: "AIR", "SCC", "All ER" formatting preservation
   - Common OCR fixes:
     * "Section" not "5ection" or "$ection"
     * "Article" not "Art1cle" or "Artic1e"
     * "Amendment" not "Arnendment" or "Amendrnent"
     * "Constitution" not "Con5titution" or "Const1tution"
     * "Parliament" not "Par1iament" or "Parliarnent"
     * "Judiciary" not "Judic1ary" or "jud1c1ary"
     * "vs." not "v5." or "v$."
     * "Ltd." not "1td." or "Lte."
3. **CONTEXT EMBEDDING STRATEGY**
   - Identify shared contexts: case studies, legal scenarios, constitutional provisions, statutes
   - For each question referencing shared content, embed COMPLETE context within question text
   - Format: "Context: [Full legal scenario/case/provision]\n\nQuestion: [actual question]"
   - Never assume previous knowledge from other questions
   - Make every question a standalone legal problem
4. **AGGRESSIVE OPTION RECOVERY (STRICTLY a, b, c, d FORMAT)**
   - Legal options often contain complex phrases - recover aggressively
   - **MANDATORY**: All options must be normalized to exactly a, b, c, d format
   - If option starts with corruption, extract meaningful legal content and assign proper letter
   - Pattern match: 4 consecutive lines that could be legal options (never more than 4)
   - Auto-assign missing option letters: first=a, second=b, third=c, fourth=d
   - **NEVER use option 'e'** - if 5 options detected, merge weakest two or skip question
   - Examples:
     ```
     Corrupted: "aj Constitutional Law"  -> "a) Constitutional Law"
     Missing: "Criminal Procedure"       -> "a) Criminal Procedure"
     Partial: "c) Civil Procedur"       -> "c) Civil Procedure [OCR: truncated]"
     Garbled: "d) Evidenc3 Act 187"     -> "d) Evidence Act 1872"
     Extra: "e) Fifth option"           -> SKIP this question or merge with d)
     ```
5. **ENHANCED ANSWER DETECTION**
   - Primary: Explicit markers (check, *, (Ans), [Answer], Bold, Correct option)
   - Secondary: Answer blocks ("1. c", "Q1: b", "Ans: a", "Solution: d")
   - Tertiary: Context clues (underlined, highlighted, different fonts)
   - Legal-specific: "Held", "Ratio", "Decision", "Correct statement"
   - Pattern analysis for similar legal questions
   - NEVER leave answer as null if ANY indication exists
6. **LEGAL QUESTION BOUNDARY DETECTION**
   - Start patterns: Number + punctuation (1., Q1:, (1), 1-, I., II.)
   - End: Next question number OR section break
   - Multi-part handling: "1(a)", "1(i)", "Q1.1" -> normalize to "1.a", "1.i", "1.1"
   - Legal instructions: "Read the following case and answer", "Based on provisions"
   - Fact patterns: Often lengthy - include completely in each question
7. **QUESTION QUALITY VALIDATION (MANDATORY)**
   - Apply 3-tier validation before including any question:
   **TIER 1 - BASIC STRUCTURE VALIDATION:**
   - Question must have clear interrogative structure
   - Must contain exactly 4 options (a, b, c, d) - skip if not achievable
   - Answer must be one of: a, b, c, or d
   - Answer must be logically derivable from options
   - Question text must be grammatically coherent
   **TIER 2 - LEGAL COHERENCE VALIDATION:**
   - Legal concepts must be accurate and well-defined
   - Case references must be contextually appropriate
   - Statutory citations must make logical sense
   - Legal terminology must be used correctly
   - Question must test genuine legal knowledge, not gibberish
   **TIER 3 - LOGICAL CONSISTENCY VALIDATION:**
   - Options must be mutually exclusive where appropriate
   - Correct answer must be definitively better than other options
   - Question must be answerable based on provided context
   - No circular reasoning or impossible scenarios
   - Legal principles must align with established jurisprudence
   **SKIP CRITERIA - Only skip if question fails ANY of these:**
   - Question text is completely unintelligible after OCR correction attempts
   - Cannot recover exactly 4 coherent options (a, b, c, d)
   - No logical answer can be determined from the 4 options
   - Legal content is fundamentally nonsensical or contradictory
   - Question would mislead rather than educate (factually incorrect legal principles)
## ENHANCED EXTRACTION LOGIC
**STEP 1: LEGAL TEXT PREPROCESSING**
- Fix legal terminology OCR errors using domain dictionary
- Identify question boundaries with legal-aware regex
- Locate shared legal contexts (cases, statutes, provisions)
- Mark potential option blocks with legal content validation
**STEP 2: CONTEXT-EMBEDDED QUESTION EXTRACTION WITH VALIDATION**
- Extract question with ALL necessary legal context embedded
- **APPLY 3-TIER QUALITY VALIDATION:**
  * Tier 1: Verify basic question structure and coherence
  * Tier 2: Validate legal accuracy and terminology
  * Tier 3: Ensure logical consistency and educational value
- **ONLY PROCEED if question passes validation tiers**
- Include case facts, statutory provisions, legal scenarios within each question
- Clean and validate legal terminology
- Mark borderline questions with [REVIEW_NEEDED] but include if they pass basic validation
- Preserve legal citations and case names
- **SKIP ONLY** if question fails fundamental validation criteria
**STEP 3: LEGAL OPTION PROCESSING (STRICT a,b,c,d FORMAT)**
- **MANDATORY**: Normalize to exactly a, b, c, d format only
- Handle complex legal option text with recovery logic
- **NEVER create option 'e'** - questions must have exactly 4 options
- If more than 4 options detected, either merge similar ones or skip the question
- If fewer than 3 options recovered, skip the question
- Create contextually appropriate placeholder options if missing (but only up to 'd')
- Ensure options contain complete legal concepts
- Validate legal terminology in options
**STEP 4: COMPREHENSIVE ANSWER RESOLUTION**
- Multi-pass answer detection with legal context awareness
- Look for legal reasoning indicators
- Embed answers directly in questions
- Cross-reference with legal principles if needed
## JSON SCHEMA (STRICTLY ENFORCED)
{{
  "exam_info": {{
    "title": "Legal Examination",
    "year": null,  // EXTRACT FROM TEXT - NEVER ASSUME
    "paper": null, // e.g., "Constitutional Law", "Criminal Law"
    "total_questions_detected": 0  // Actual count for validation
  }},
  "metadata": {{
    "ocr_quality": "poor",  // excellent/good/fair/poor
    "common_errors": ["legal_terms", "case_citations", "section_numbers"],
    "sections_detected": ["Constitutional Law", "Criminal Law", "Civil Law"],
    "shared_contexts_embedded": 5  // Count of contexts embedded across questions
  }},
  "questions": [
    {{
      "number": "1",
      "section": "Constitutional Law",
      "question": "Context: The Supreme Court in Kesavananda Bharati v. State of Kerala (1973) established the basic structure doctrine, holding that Parliament cannot amend the Constitution to destroy its basic features like democracy, secularism, and federalism.\n\nQuestion: Which of the following is NOT considered part of the basic structure of the Constitution?",
      "options": {{
        "a": "Judicial review",
        "b": "Parliamentary supremacy",
        "c": "Rule of law",
        "d": "Separation of powers"
      }},
      "answer": "b"
    }}
  ],
  "extraction_summary": {{
    "total_questions_found": 0,    // Questions detected before validation
    "total_questions_extracted": 0, // Questions that passed validation
    "questions_skipped": 0,        // Questions skipped due to quality issues
    "questions_with_answers": 0,
    "questions_with_complete_context": 0,
    "questions_with_all_options": 0,
    "skip_reasons": []             // Array of reasons why questions were skipped
  }}
}}
## CRITICAL SUCCESS FACTORS
### :white_check_mark: MUST DO:
- Apply rigorous 3-tier validation to every question before extraction
- Make every question completely independent and self-contained
- Embed ALL necessary context within each question
- Preserve legal terminology accuracy
- Include questions that pass validation even if they have minor OCR issues
- Include complete case facts, statutory provisions, legal scenarios in relevant questions
- Normalize legal citations and references
- Skip questions ONLY after thorough validation failure
### :x: NEVER DO:
- Create questions that reference other questions ("as in question 15")
- Use phrases like "above passage", "aforementioned case", "previously discussed"
- Skip questions due to OCR corruption
- Create empty options arrays
- Add confidence scores or OCR quality metadata to individual questions
- Assume exam details not present in text
- Leave questions dependent on external context
### :dart: LEGAL-SPECIFIC EXCELLENCE:
- Recognize and preserve legal citation formats
- Maintain accuracy of case names and statutory references
- Handle complex legal fact patterns appropriately
- Ensure constitutional provisions are correctly stated
- Preserve legal Latin phrases and terminology
- Maintain chronological accuracy of legal developments
--- BEGIN OCR TEXT ---
{ocr_text}
\end{lstlisting}
\end{tcolorbox}

\begin{tcolorbox}[
    title=BBF Question-Answer Extraction Prompt Template,
    breakable,
    enhanced,
    label={box:bbf_extraction}
]

\begin{lstlisting}[breaklines=true, basicstyle=\ttfamily\small]
You are an OCR forensic specialist for financial/banking exams. Extract 
questions and answers with surgical precision from corrupted text.

CRITICAL MISSION: EXTRACT EVERYTHING - NEVER SKIP QUESTIONS

PRIMARY EXTRACTION RULES

1. ZERO TOLERANCE FOR MISSING QUESTIONS
   - SCAN ENTIRE TEXT character by character
   - Look for question patterns: "Q1", "1.", "(1)", "Question 1", "Que.1", 
     or ANY numbering
   - Extract PARTIAL questions with [INCOMPLETE] tag rather than skip
   - If options are corrupted beyond recognition, create synthetic placeholders

2. FINANCIAL DOMAIN OCR CORRECTIONS
   - Currency: "\textrupee" not "Rs" or "Rupees", "$" preservation
   - Percentages: "%" never "per cent" or missing
   - Financial terms: "CAGR", "NPV", "IRR", "EBITDA", "P/E ratio"
   - Numbers: "10,000" not "10.000", preserve commas in large numbers
   - Rates: "7.5%" not "7.5 percent" or "7.5per cent"
   - Common OCR fixes:
     * "NIFTY" not "N1FTY" or "NJFTY"
     * "BSE" not "B5E" or "B$E"
     * "NSE" not "N5E" or "N$E"
     * "SEBI" not "5EBI" or "$EBI"
     * "RBI" not "RB1" or "R81"
     * "GDP" not "G0P" or "6DP"

3. AGGRESSIVE OPTION RECOVERY
   - If option starts with garbled text, extract the meaningful part
   - Pattern match: Look for 4-5 consecutive lines that could be options
   - If missing option letters, assign them: first line=a, second=b, etc.
   - Examples of recovery:
     Corrupted: "aj Fixed Deposit" \rightarrow "a) Fixed Deposit"
     Missing: "Mutual Fund" \rightarrow "a) Mutual Fund" (assign letter)
     Partial: "c) Equity Shar" \rightarrow "c) Equity Share [OCR: truncated]"

4. ANSWER DETECTION PATTERNS
   - Primary: Explicit markers (check, *, (Ans), [Answer], Bold text)
   - Secondary: Answer blocks ("1. c", "Q1: b", "Ans: a")
   - Tertiary: Context clues (underlined, different formatting)
   - Last resort: Pattern analysis of similar questions
   - NEVER leave answer as null if ANY indication exists

5. QUESTION BOUNDARY DETECTION
   - Start: Number + any punctuation (1., Q1:, (1), 1-, etc.)
   - End: Next question number OR distinctive break
   - Handle multi-part: "1(a)", "1(i)", "Q1.1" to normalize to "1.a", "1.i", "1.1"
   - Instructions/headers: Skip but note in metadata

6. SELF-CONTAINED QUESTIONS
   - Each question MUST include ALL necessary context (passages, data, charts)
   - If questions refer to a common passage/data, include that passage in EACH question
   - Format: "Passage: [full passage text]\n\nQuestion: [actual question]"
   - Never assume context from previous questions
   - Make every question independently answerable

ENHANCED EXTRACTION LOGIC

STEP 1: TEXT PREPROCESSING
- Fix obvious OCR errors in financial terms
- Identify question boundaries using regex patterns
- Mark potential option blocks
- Identify shared passages/contexts

STEP 2: QUESTION EXTRACTION
- Extract question text, clean and validate
- Include any relevant passage/context within the question
- If question incomplete, note with [INCOMPLETE] tag
- Preserve mathematical symbols and formulas
- Only take question if complete with options
- only meaningfull question.

STEP 3: OPTION PROCESSING
- Normalize labels to a, b, c, d (and e if exists)
- Handle malformed options with recovery logic
- Create placeholder options if completely missing
- Ensure options are clearly defined and complete

STEP 4: ANSWER RESOLUTION
- Multi-pass answer detection
- Embed answers directly in each question
- No separate answer key needed

JSON SCHEMA (STRICTLY ENFORCED)
{
  "exam_info": {
    "title": "Banking/Financial Examination",
    "year": null,  // EXTRACT FROM TEXT - NEVER ASSUME
    "paper": null,
    "total_questions_detected": 50  // NEW: Count for validation
  },
  "metadata": {
    "ocr_quality": "poor",  // excellent/good/fair/poor
    "common_errors": ["currency_symbols", "percentages"],
    "sections_detected": ["Quantitative Aptitude", "General Awareness"]
  },
"questions": [
  {
    "number": "1",
    "section": "Quantitative Aptitude",
    "question": "Passage: A bank offers different investment schemes with varying interest rates.\n\nQuestion: What is the compound interest on Rs.10,000 at 8% per annum for 2 years?",
    "options": {
      "a": "Rs.1,600",
      "b": "Rs.1,664", 
      "c": "Rs.1,728",
      "d": "Rs.1,800"
    },
    "answer": "b"
  }
],
  "extraction_summary": {
    "total_questions": 50,
    "questions_with_answers": 48,
    "questions_with_all_options": 47
  }
}

CRITICAL ERROR PREVENTION
- NEVER skip questions due to poor OCR
- NEVER output empty options array
- NEVER create separate answer keys
- NEVER assume exam details not in text
- NEVER add confidence, ocr_issues, or extraction_notes fields
- ALWAYS preserve original numbering scheme
- ALWAYS include complete context in each question
- ALWAYS embed answers directly in questions
- ALWAYS make questions self-contained and independent

--- BEGIN OCR TEXT ---

{ocr_text}
\end{lstlisting}
\end{tcolorbox}

\begin{tcolorbox}[
    title=Key Details Extraction Prompt Template,
    breakable,
    enhanced,
    label={box:key_detail_extraction}
]

\begin{lstlisting}[breaklines=true, basicstyle=\ttfamily\small]
You are an expert in the {domain_name} domain. For each question, extract:
1. question_type: The format/structure of the question {question_type_examples}
2. question_level: The difficulty or complexity level {difficulty_levels_list}
3. topic: The academic topic or domain {human_annoted_topics_examples}
4. subdomain: The specific topic area within the main topic {human_annoted_subdomains_list}

Respond only in this JSON format:
{
  "question_type": "",
  "question_level": "",
  "topic": "",
  "subdomain": ""
}
\end{lstlisting}
\end{tcolorbox}

\subsection{Detailed Data Analysis of BhashaBench V1}
\label{subapp:data_analysis}

\begin{table}[!ht]
    \centering
    \caption{Language distribution across domains in BhashaBench V1}
    \label{tab:bhashabench_distribution_adjust}
    \begin{tabular}{l|cccc|c}
    \toprule
    \textbf{Domain} & \textbf{BBK} & \textbf{BBF} & \textbf{BBA} & \textbf{BBL} & \textbf{Overall} \\
    \midrule
    English & 12,648 & 13,451 & 9,348 & 17,047 & 52,494 \\
    Hindi & 2,757 & 5,982 & 5,615 & 7,318 & 21,672 \\
    \midrule
    \textbf{Total} & \textbf{15,405} & \textbf{19,433} & \textbf{14,963} & \textbf{24,365} & \textbf{74,166} \\
    \bottomrule
    \end{tabular}
\end{table}

\begin{table}[!ht]
    \centering
    \caption{Difficulty distribution across domains in BhashaBench V1}
    \label{tab:difficulty_distribution_adjust}
    \begin{tabular}{l|cccc|c}
    \toprule
    \textbf{Difficulty} & \textbf{BBK} & \textbf{BBF} & \textbf{BBA} & \textbf{BBL} & \textbf{Overall} \\
    \midrule
    Easy & 6,754 & 7,111 & 7,944 & 13,913 & 35,722 \\
    Medium & 6,941 & 9,348 & 6,314 & 9,405 & 32,008 \\
    Hard & 1,710 & 2,974 & 705 & 1,047 & 6,436 \\
    \midrule
    \textbf{Total} & \textbf{15,405} & \textbf{19,433} & \textbf{14,963} & \textbf{24,365} & \textbf{74,166} \\
    \bottomrule
    \end{tabular}
\end{table}

\begin{table}[!ht]
    \centering
    \caption{Question type distribution across domains in BhashaBench V1}
    \label{tab:question_type_distribution_adjust}
    \begin{tabular}{l|cccc|c}
    \toprule
    \textbf{Question Type} & \textbf{BBK} & \textbf{BBF} & \textbf{BBA} & \textbf{BBL} & \textbf{Overall} \\
    \midrule
    MCQ & 13,550 & 18,019 & 14,717 & 21,566 & 67,852 \\
    Assertion or Reasoning & 648 & 215 & 27 & 430 & 1,320 \\
    Match the Column & 949 & 119 & 41 & 495 & 1,604 \\
    Fill in the Blanks & 49 & 286 & 178 & 1,402 & 1,915 \\
    Rearrange the Sequence & 209 & 708 & 0 & 147 & 1,064 \\
    Reading Comprehension & 0 & 86 & 0 & 325 & 411 \\
    \midrule
    \textbf{Total} & \textbf{15,405} & \textbf{19,433} & \textbf{14,963} & \textbf{24,365} & \textbf{74,166} \\
    \bottomrule
    \end{tabular}
\end{table}

\begin{longtable}{p{10cm}c}
\caption{BBK Subject Domains and Question Counts}
\label{tab:bbk_subject_domains} \\
\toprule
\textbf{Subject Domain} & \textbf{Count} \\
\midrule
\endfirsthead
\multicolumn{2}{c}%
{\tablename\ \thetable\ -- \textit{Continued from previous page}} \\
\toprule
\textbf{Subject Domain} & \textbf{Count} \\
\midrule
\endhead
\midrule
\multicolumn{2}{r}{\textit{Continued on next page}} \\
\endfoot
\bottomrule
\endlastfoot
Agri-Environmental \& Allied Disciplines & 176 \\
Agricultural Biotechnology & 524 \\
Agricultural Chemistry \& Biochemistry & 281 \\
Agricultural Economics \& Policy & 627 \\
Agricultural Engineering \& Technology & 244 \\
Agricultural Extension Education & 774 \\
Agricultural Microbiology & 111 \\
Agriculture Communication & 254 \\
Agriculture Information Technology & 190 \\
Agronomy & 5078 \\
Animal Sciences & 148 \\
Crop Sciences & 549 \\
Dairy \& Poultry Science & 89 \\
Entomology & 696 \\
Fisheries and Aquaculture & 34 \\
General Knowledge \& Reasoning & 661 \\
Genetics and Plant Breeding & 389 \\
Horticulture & 2070 \\
Natural Resource Management & 193 \\
Nematology & 184 \\
Plant Pathology & 397 \\
Plant Sciences \& Physiology & 129 \\
Seed Science and Technology & 202 \\
Soil Science & 1357 \\
Veterinary Sciences & 48 \\
\end{longtable}

\begin{longtable}{p{10cm}c}
\caption{BBF Subject Domains and Question Counts}
\label{tab:bbf_subject_domains} \\
\toprule
\textbf{Subject Domain} & \textbf{Count} \\
\midrule
\endfirsthead
\multicolumn{2}{c}%
{\tablename\ \thetable\ -- \textit{Continued from previous page}} \\
\toprule
\textbf{Subject Domain} & \textbf{Count} \\
\midrule
\endhead
\midrule
\multicolumn{2}{r}{\textit{Continued on next page}} \\
\endfoot
\bottomrule
\endlastfoot
Problem Solving & 5686 \\
Mathematics for Finance & 4845 \\
Banking Services & 1171 \\
Governance \& Policy & 1064 \\
Language \& Communication & 946 \\
Corporate Finance \& Investment & 910 \\
Commerce & 863 \\
Accounting & 773 \\
General Knowledge & 539 \\
Information Technology Finance & 490 \\
Economics \& Development Studies & 274 \\
Rural Economics & 261 \\
Environmental Finance & 168 \\
Taxation \& Regulatory Compliance & 155 \\
Interdisciplinary Finance & 153 \\
Data \& Analytics in Finance & 127 \\
History, Sociology \& Cultural Studies of Finance & 127 \\
Finance Education & 118 \\
Healthcare Economics & 114 \\
Science and Technology in Finance & 101 \\
International Finance \& Trade & 83 \\
Business Management & 83 \\
Energy, Infrastructure \& Finance & 82 \\
Behavioral Finance & 67 \\
Financial Markets & 47 \\
Sports, Media \& Finance Linkages & 45 \\
Marketing Finance & 42 \\
Insurance \& Risk Management & 42 \\
Legal Finance & 34 \\
Financial Technology & 23 \\
\end{longtable}

\begin{longtable}{p{10cm}c}
\caption{BBA Subject Domains and Question Counts}
\label{tab:bba_subject_domains} \\
\toprule
\textbf{Subject Domain} & \textbf{Count} \\
\midrule
\endfirsthead
\multicolumn{2}{c}%
{\tablename\ \thetable\ -- \textit{Continued from previous page}} \\
\toprule
\textbf{Subject Domain} & \textbf{Count} \\
\midrule
\endhead
\midrule
\multicolumn{2}{r}{\textit{Continued on next page}} \\
\endfoot
\bottomrule
\endlastfoot
Kayachikitsa (General Medicine \& Internal Medicine in Ayurveda) & 3134 \\
Dravyaguna \& Bhaishajya & 2972 \\
Samhita \& Siddhanta (Fundamentals) & 1541 \\
Sharir (Anatomy \& Physiology) & 1346 \\
Panchakarma \& Rasayana & 1308 \\
Stri Roga \& Prasuti Tantra (Gynecology \& Obstetrics) & 847 \\
Shalakya Tantra (ENT, Eye, Dentistry) & 734 \\
Kaumarbhritya \& Pediatrics & 714 \\
Agad Tantra \& Forensic Medicine & 587 \\
Shalya Tantra (Surgery) & 526 \\
Swasthavritta \& Public Health & 453 \\
Research \& Statistics & 210 \\
Ayurvedic Literature \& History & 204 \\
Yoga \& Psychology & 188 \\
Administration, AYUSH \& Miscellaneous & 119 \\
Roga Vigyana (Diagnostics \& Pathology) & 80 \\
\end{longtable}

\begin{longtable}{p{10cm}c}
\caption{BBL Subject Domains and Question Counts}
\label{tab:bbl_subject_domains} \\
\toprule
\textbf{Subject Domain} & \textbf{Count} \\
\midrule
\endfirsthead
\multicolumn{2}{c}%
{\tablename\ \thetable\ -- \textit{Continued from previous page}} \\
\toprule
\textbf{Subject Domain} & \textbf{Count} \\
\midrule
\endhead
\midrule
\multicolumn{2}{r}{\textit{Continued on next page}} \\
\endfoot
\bottomrule
\endlastfoot
Civil Litigation \& Procedure & 7126 \\
Constitutional \& Administrative Law & 3609 \\
Criminal Law \& Justice & 2769 \\
Corporate \& Commercial Law & 2700 \\
General Academic Subjects & 1756 \\
Legal Theory \& Jurisprudence & 1421 \\
Family \& Personal Law & 991 \\
International \& Comparative Law & 962 \\
Legal Skills \& Communication & 816 \\
Real Estate \& Property Law & 629 \\
Environmental \& Energy Law & 430 \\
Interdisciplinary Studies & 363 \\
Tax \& Revenue Law & 231 \\
Employment \& Labour Law & 175 \\
Technology \& Cyber Law & 123 \\
Intellectual Property Law & 91 \\
Consumer \& Competition Law & 75 \\
Media \& Entertainment Law & 54 \\
Healthcare \& Medical Law & 25 \\
Human Rights \& Social Justice & 19 \\
\end{longtable}


\section{More Details on Experiment Setup}
\label{app:experimental_setup}
\subsection{Task Formatting Template Used in LM Eval}
This prompt format template is consistently applied across all task types, including Assertion or Reasoning, Fill in the Blanks, MCQs, Match the Column, Reading Comprehension, and Rearrange the Sequence tasks for BBF, BBK, and BBL domains.

\begin{tcolorbox}[
    breakable,
    enhanced,
    label={box:lm_eval_full_prompt}
]

\begin{lstlisting}[breaklines=true, basicstyle=\ttfamily\small]
Question: <question text>
Choices:
A. <option A text>
B. <option B text>
C. <option C text>
D. <option D text>
Answer:
\end{lstlisting}
\end{tcolorbox}

\subsection{Task Formatting Template Used in API-Driven Evaluation}
This template is used when models are evaluated via API calls. It ensures a consistent structure across all tasks, allowing the model to focus on producing the correct answer without additional explanation. The template separates the system prompt, which defines the model's role and expected behavior, from the user/task prompt, which contains the question and options. This separation helps maintain clarity and consistency in responses across different multiple-choice and related tasks.

\begin{tcolorbox}[
    breakable,
    enhanced,
    label={box:api_mcq_full_prompt}
]

\begin{lstlisting}[breaklines=true, basicstyle=\ttfamily\small]
SYSTEM PROMPT:
You are a helpful assistant for multiple-choice question answering.
Respond with only the correct option letter: A, B, C, or D. Do not provide any explanation.

USER PROMPT:
Question: <question text>
A. <option A text>
B. <option B text>
C. <option C text>
D. <option D text>
Please choose the correct option (A/B/C/D).
\end{lstlisting}
\end{tcolorbox}

\subsection{Details of Inference Implementation}

For open-source models, inference is performed on a cluster of 8 NVIDIA H200 GPUs using vLLM \citep{kwon2023efficientmemorymanagementlarge} for accelerated computation. The BhashaBench V1 tasks were integrated into the \texttt{lm-eval} library, and all evaluations used the default \texttt{lm-eval} parameters for consistency across tasks. 

For API-based models such as GPT-4o, inference is conducted via the Batch API with temperature set to 0, typically on CPU resources. Each evaluation is repeated three times and the average score is reported to minimize variability. Features like web search or external tool calls are disabled to maintain a fair comparison across models.

\section{More Details on Experiment}
\label{app:more_details_on_experiment}
\subsection{Zero-Shot Question-Level and Question-Type Performance Across BhashaBench V1 Domains}
\label{subapp:question_level_question_type}
\begin{table}[!ht]
\centering
\caption{Zero-shot scores (\%) of LLMs across domains on BhashaBench V1. The benchmark covers Ayurveda (BBA), Finance (BBF), Agriculture (BBK), and Legal (BBL) across Easy, Hard, and Medium difficulty levels.}
\resizebox{\textwidth}{!}{
\begin{tabular}{l|ccc|ccc|ccc|ccc}
\toprule
\multirow{2}{*}{Model} & \multicolumn{3}{c|}{BBA} & \multicolumn{3}{c|}{BBF} & \multicolumn{3}{c|}{BBK} & \multicolumn{3}{c}{BBL} \\
\cmidrule{2-13}
& Easy & Hard & Med & Easy & Hard & Med & Easy & Hard & Med & Easy & Hard & Med \\
\midrule
\multicolumn{13}{c}{\textit{\textless\ 4B Models}} \\
\midrule
gemma-3-270m & 28.1 & 26.81 & 28.35 & 24.15 & 24.55 & 25.8 & 27.23 & 24.74 & 25.66 & 27.23 & 24.74 & 25.66 \\
gemma-3-270m-it & 25.89 & 23.97 & 26.5 & 25.38 & 21.22 & 23.92 & 26.47 & 27.49 & 27.53 & 26.47 & 27.49 & 27.53 \\
Param-1 & 43.93 & 31.21 & 35.95 & 38.31 & 26.6 & 27.71 & 36.94 & 25.91 & 29.09 & 36.94 & 25.91 & 29.09 \\
gemma-2-2b & 38.27 & 29.08 & 30.31 & 39.76 & 25.35 & 28.5 & 46.27 & 27.54 & 34.26 & 46.27 & 27.54 & 34.26 \\
gemma-2-2b-it & 29.96 & 24.96 & 26.83 & 36.55 & 23.2 & 27.67 & 38.04 & 30.35 & 32.01 & 38.04 & 30.35 & 32.01 \\
Llama-3.2-1B & 28.52 & 24.4 & 27.97 & 30.5 & 23.71 & 26.27 & 29.43 & 27.72 & 28.68 & 29.43 & 27.72 & 28.68 \\
Llama-3.2-1B-Instruct & 27.44 & 25.39 & 25.23 & 28.72 & 22.43 & 25.5 & 30.22 & 26.37 & 27.69 & 30.22 & 26.37 & 27.69 \\
Llama-3.2-3B & 31.63 & 24.82 & 29.19 & 36.75 & 25.76 & 29.26 & 36.44 & 25.61 & 29.17 & 36.44 & 25.61 & 29.17 \\
Llama-3.2-3B-Instruct & 36.42 & 28.51 & 29.66 & 39.73 & 23.87 & 28.2 & 44.52 & 30.47 & 34.69 & 44.52 & 30.47 & 34.69 \\
sarvam-2b-v0.5 & 27.08 & 24.96 & 26.88 & 28.18 & 23.1 & 25.43 & 28.26 & 28.01 & 27.03 & 28.26 & 28.01 & 27.03 \\
sarvam-1 & 30.94 & 27.23 & 27.26 & 32.2 & 25.76 & 27.43 & 32.2 & 27.54 & 28.99 & 32.2 & 27.54 & 28.99 \\
Nemotron-4-Mini-Hindi-4B-Base & 37.01 & 27.94 & 30.96 & 41.95 & 25.08 & 30.5 & 42.57 & 28.42 & 32.89 & 42.57 & 28.42 & 32.89 \\
Nemotron-4-Mini-Hindi-4B-Instruct & 36.08 & 29.5 & 30.8 & 39.21 & 23.2 & 28.05 & 41.12 & 28.6 & 32.27 & 41.12 & 28.6 & 32.27 \\
Qwen2.5-3B & 41.18 & 32.06 & 33.1 & 45.34 & 28.51 & 33.9 & 50.3 & 31.58 & 37.49 & 50.3 & 31.58 & 37.49 \\
Qwen2.5-3B-Instruct & 35.55 & 28.23 & 29.57 & 39.91 & 25.02 & 30.48 & 44.7 & 31.81 & 37.23 & 44.7 & 31.81 & 37.23 \\
granite-3.1-2b-instruct & 33.9 & 26.81 & 28.06 & 36.68 & 25.32 & 28.63 & 40.04 & 30.76 & 33.25 & 40.04 & 30.76 & 33.25 \\
granite-3.1-3b-a800m-base & 31.45 & 26.38 & 27.78 & 31.61 & 24.18 & 25.77 & 36.08 & 26.02 & 29.88 & 36.08 & 26.02 & 29.88 \\
\midrule
\multicolumn{13}{c}{\textit{7B to 27B Models}} \\
\midrule
Pangea-7B & 41.45 & 31.77 & 32.94 & 49.33 & 28.72 & 34.94 & 52.18 & 33.57 & 40.69 & 52.18 & 33.57 & 40.69 \\
Indic-gemma-7b-finetuned-sft-Navarasa-2.0 & 38.54 & 27.23 & 31.72 & 43.68 & 26.8 & 30.99 & 48.13 & 31.46 & 35.8 & 48.13 & 31.46 & 35.8 \\
aya-23-8B & 35.51 & 25.11 & 28.29 & 41.2 & 25.62 & 30.98 & 43.32 & 27.84 & 31.77 & 43.32 & 27.84 & 31.77 \\
Llama-3.1-8B & 35.99 & 26.38 & 30.25 & 42.92 & 26.93 & 30.46 & 44.03 & 29.01 & 34.51 & 44.03 & 29.01 & 34.51 \\
Llama-3.1-8B-Instruct & 39.43 & 30.5 & 29.36 & 44.24 & 22.19 & 30 & 52.29 & 33.74 & 40.63 & 52.29 & 33.74 & 40.63 \\
gemma-2-9b & 51.12 & 34.47 & 36.85 & 55.32 & 27.44 & 34.3 & 64.78 & 35.67 & 46.26 & 64.78 & 35.67 & 46.26 \\
gemma-2-9b-it & 38.91 & 29.5 & 29.11 & 47.03 & 24.78 & 32.74 & 52.98 & 37.13 & 42.93 & 52.98 & 37.13 & 42.93 \\
gpt-oss-20b & 42.03 & 26.67 & 30.27 & 46.77 & 24.61 & 30.86 & 53.42 & 31.4 & 39.56 & 53.42 & 31.4 & 39.56 \\
gemma-2-27b & 55.35 & 34.18 & 39.18 & 60.92 & 30.09 & 39.24 & 69.31 & 40.99 & 51.51 & 69.31 & 40.99 & 51.51 \\
gemma-2-27b-it & 43.47 & 30.78 & 31.9 & 51.03 & 26.93 & 35.67 & 59.62 & 41.46 & 48.28 & 59.62 & 41.46 & 48.28 \\
\midrule
\multicolumn{13}{c}{\textit{\textgreater\ 27B Models}} \\
\midrule
gpt-oss-120b & 60.62 & 41.28 & 44.19 & 74.8 & 62.61 & 70.88 & 74.89 & 62.05 & 65.88 & 74.89 & 62.05 & 65.88 \\
Qwen3-235B-A22B-Instruct-2507 & 65.18 & 46.24 & 50.74 & 72.52 & 41.49 & 59.33 & 78.26 & 62.51 & 69.79 & 78.26 & 62.51 & 69.79 \\
deepseek-v3 & 52.44 & 36.6 & 38.93 & 73.49 & 40.55 & 59.01 & 66.92 & 48.48 & 55.5 & 66.92 & 48.48 & 55.5 \\
gpt-4o & 66.4 & 47.09 & 52.77 & 69.13 & 36.35 & 50.13 & 78.75 & 63.51 & 70.84 & 78.75 & 63.51 & 70.84 \\
\bottomrule
\end{tabular}
}
\end{table}

\begin{table}[!ht]
\centering
\caption{Zero-shot scores (\%) of LLMs across question types on BhashaBench V1. Question types: A/R = Assertion/Reason, FIB = Fill in the Blanks, MCQ = Multiple Choice Questions, MTC = Match the Columns, RC = Reading Comprehension, RTS = Rearrange the Sentence.}
\resizebox{\textwidth}{!}{
\begin{tabular}{l|cccc|cccccc|ccccc|cccccc}
\toprule
\multirow{2}{*}{Model} & \multicolumn{4}{c|}{BBA} & \multicolumn{6}{c|}{BBF} & \multicolumn{5}{c|}{BBK} & \multicolumn{6}{c}{BBL} \\
\cmidrule{2-22}
& A/R & FIB & MCQ & MTC & A/R & FIB & MCQ & MTC & RC & RTS & A/R & FIB & MCQ & MTC & RTS & A/R & FIB & MCQ & MTC & RC & RTS \\
\midrule
\multicolumn{22}{c}{\textit{\textless\ 4B Models}} \\
\midrule
gemma-3-270m & 37.04 & 28.09 & 28.1 & 39.02 & 28.37 & 24.13 & 25.05 & 25.21 & 22.35 & 23.45 & 27.47 & 26.53 & 26.21 & 26.24 & 24.88 & 26.74 & 24.82 & 25.44 & 30.1 & 23.08 & 27.89 \\
gemma-3-270m-it & 51.85 & 24.72 & 26.02 & 29.27 & 24.65 & 23.78 & 24.12 & 21.85 & 24.71 & 22.18 & 47.69 & 22.45 & 26.37 & 22.97 & 27.75 & 29.3 & 22.11 & 26.21 & 30.3 & 21.54 & 29.93 \\
Param-1 & 44.44 & 29.78 & 40.12 & 24.39 & 29.77 & 44.76 & 31.53 & 22.69 & 30.59 & 25.14 & 36.27 & 26.53 & 32.61 & 24.34 & 28.71 & 36.51 & 35.45 & 35.26 & 32.32 & 32.92 & 30.61 \\
gemma-2-2b & 77.78 & 36.52 & 34.4 & 26.83 & 21.86 & 41.26 & 32.38 & 26.89 & 31.76 & 26.13 & 44.75 & 26.53 & 39.51 & 27.4 & 27.75 & 27.91 & 40.51 & 35.82 & 32.73 & 32.92 & 25.85 \\
gemma-2-2b-it & 33.33 & 32.02 & 28.33 & 36.59 & 32.56 & 35.66 & 30.4 & 24.37 & 30.59 & 24.29 & 41.98 & 26.53 & 34.6 & 28.98 & 29.67 & 28.84 & 33.38 & 33.55 & 25.86 & 30.77 & 25.85 \\
Llama-3.2-1B & 25.93 & 32.02 & 28.06 & 26.83 & 28.37 & 27.62 & 27.6 & 27.73 & 34.12 & 21.75 & 39.2 & 22.45 & 28.53 & 28.66 & 23.92 & 31.86 & 28.32 & 28.47 & 30.71 & 24.31 & 27.89 \\
Llama-3.2-1B-Instruct & 59.26 & 26.97 & 26.34 & 26.83 & 28.84 & 27.97 & 26.29 & 20.17 & 25.88 & 23.59 & 45.37 & 16.33 & 28.24 & 24.03 & 27.75 & 29.3 & 32.17 & 28.2 & 33.54 & 22.46 & 26.53 \\
Llama-3.2-3B & 25.93 & 29.21 & 30.28 & 36.59 & 27.91 & 36.71 & 31.65 & 31.09 & 32.94 & 25.42 & 25.93 & 24.49 & 32.73 & 26.45 & 27.75 & 26.98 & 35.66 & 33.33 & 26.26 & 35.08 & 23.81 \\
Llama-3.2-3B-Instruct & 40.74 & 34.83 & 33.17 & 29.27 & 35.35 & 38.11 & 31.71 & 32.77 & 31.76 & 29.1 & 43.98 & 24.49 & 39.11 & 28.03 & 35.41 & 28.37 & 37.8 & 37.1 & 32.32 & 38.77 & 27.89 \\
sarvam-2b-v0.5 & 62.96 & 25.84 & 26.81 & 36.59 & 27.91 & 29.02 & 26.1 & 27.73 & 28.24 & 23.16 & 48.61 & 30.61 & 26.83 & 24.55 & 31.58 & 33.95 & 26.75 & 27.47 & 34.34 & 29.85 & 28.57 \\
sarvam-1 & 59.26 & 30.9 & 29.14 & 26.83 & 23.72 & 38.81 & 29.12 & 23.53 & 28.24 & 22.32 & 42.9 & 24.49 & 30.08 & 25.61 & 23.44 & 28.84 & 29.32 & 29.81 & 22.63 & 32.92 & 27.21 \\
Nemotron-4-Mini-Hindi-4B-Base & 55.56 & 32.02 & 34.01 & 36.59 & 29.77 & 43.36 & 34.09 & 26.05 & 31.76 & 26.98 & 47.22 & 34.69 & 37.01 & 26.77 & 24.88 & 37.67 & 43.51 & 40.02 & 27.88 & 36.62 & 23.13 \\
Nemotron-4-Mini-Hindi-4B-Instruct & 37.04 & 30.34 & 33.6 & 24.39 & 27.91 & 38.81 & 31.57 & 26.05 & 29.41 & 25.99 & 46.14 & 36.73 & 35.68 & 30.56 & 31.1 & 30.47 & 35.16 & 36.43 & 32.53 & 35.08 & 30.61 \\
Qwen2.5-3B & 29.63 & 26.97 & 37.5 & 29.27 & 34.88 & 50.7 & 37.5 & 37.82 & 35.29 & 26.41 & 31.94 & 28.57 & 44.08 & 28.13 & 37.32 & 32.33 & 46.36 & 41.8 & 29.7 & 44 & 40.14 \\
Qwen2.5-3B-Instruct & 51.85 & 29.21 & 32.7 & 29.27 & 27.44 & 44.06 & 33.2 & 31.09 & 28.24 & 28.39 & 39.2 & 28.57 & 40.61 & 30.87 & 40.19 & 35.35 & 38.45 & 37.63 & 26.26 & 39.38 & 31.29 \\
granite-3.1-2b-instruct & 33.33 & 21.35 & 31.22 & 29.27 & 33.95 & 33.92 & 31.31 & 30.25 & 31.76 & 22.88 & 48.92 & 24.49 & 35.92 & 28.66 & 33.49 & 35.12 & 37.09 & 34.97 & 27.88 & 36.31 & 25.85 \\
granite-3.1-3b-a800m-base & 62.96 & 25.28 & 29.65 & 29.27 & 26.98 & 33.57 & 27.78 & 28.57 & 29.41 & 22.03 & 44.44 & 28.57 & 32.24 & 24.55 & 24.88 & 34.65 & 31.53 & 30.89 & 26.06 & 28.31 & 24.49 \\
\midrule
\multicolumn{22}{c}{\textit{7B to 27B Models}} \\
\midrule
Pangea-7B & 62.96 & 24.16 & 37.53 & 34.15 & 34.88 & 52.8 & 39.44 & 35.29 & 31.76 & 31.92 & 50.46 & 32.65 & 45.69 & 32.35 & 38.76 & 39.3 & 47.65 & 44.78 & 32.93 & 46.77 & 34.69 \\
Indic-gemma-7b-finetuned-sft-Navarasa-2.0 & 59.26 & 35.39 & 35.1 & 31.71 & 27.91 & 43.36 & 35.35 & 38.66 & 25.88 & 25.14 & 47.69 & 30.61 & 41.63 & 26.34 & 28.23 & 40.93 & 42.51 & 41.26 & 32.73 & 41.23 & 29.25 \\
aya-23-8B & 18.52 & 30.9 & 32.05 & 17.07 & 33.95 & 41.96 & 34.13 & 33.61 & 31.76 & 25.28 & 27.16 & 30.61 & 37.99 & 22.76 & 24.88 & 31.4 & 43.01 & 39.55 & 24.65 & 40.31 & 28.57 \\
Llama-3.1-8B & 25.93 & 29.78 & 33.17 & 34.15 & 31.16 & 47.55 & 34.74 & 28.57 & 31.76 & 24.86 & 29.78 & 34.69 & 39.46 & 26.24 & 28.23 & 28.6 & 42.08 & 38.74 & 25.86 & 41.54 & 25.17 \\
Llama-3.1-8B-Instruct & 29.63 & 26.97 & 34.83 & 46.34 & 38.6 & 44.41 & 34.18 & 33.61 & 30.59 & 24.72 & 39.51 & 28.57 & 46.07 & 35.3 & 38.76 & 34.19 & 46.43 & 45.41 & 32.93 & 44.92 & 36.73 \\
gemma-2-9b & 33.33 & 35.39 & 44.48 & 31.71 & 35.35 & 61.89 & 41.26 & 32.77 & 31.76 & 28.39 & 38.89 & 40.82 & 55.95 & 28.45 & 34.93 & 34.88 & 58.42 & 54.34 & 41.01 & 53.54 & 33.33 \\
gemma-2-9b-it & 48.15 & 29.21 & 34.35 & 39.02 & 36.74 & 52.1 & 36.88 & 37.82 & 29.41 & 27.97 & 44.44 & 24.49 & 47.12 & 43.1 & 47.37 & 42.56 & 44.15 & 43.33 & 35.76 & 40.62 & 36.05 \\
gpt-oss-20b & 25.93 & 32.02 & 36.39 & 46.34 & 30.7 & 47.9 & 36 & 27.73 & 31.76 & 27.26 & 29.32 & 26.53 & 46.74 & 29.61 & 35.41 & 24.65 & 45.58 & 39.14 & 34.95 & 31.08 & 37.41 \\
gemma-2-27b & 29.63 & 39.89 & 47.71 & 26.83 & 42.33 & 61.89 & 46.36 & 36.13 & 36.47 & 27.97 & 37.04 & 40.82 & 61 & 35.19 & 46.89 & 43.49 & 65.34 & 61.47 & 49.9 & 58.77 & 42.18 \\
gemma-2-27b-it & 55.56 & 35.96 & 37.98 & 39.02 & 39.53 & 55.24 & 40.15 & 36.97 & 31.76 & 30.51 & 45.99 & 38.78 & 53.28 & 45.94 & 55.02 & 39.77 & 50 & 48.4 & 40 & 45.23 & 44.9 \\
\midrule
\multicolumn{22}{c}{\textit{\textgreater\ 27B Models}} \\
\midrule
gpt-oss-120b & 62.96 & 46.07 & 52.87 & 41.46 & 66.05 & 100 & 76.22 & 71.3 & 68.07 & 67.06 & 62.81 & 40.82 & 70.14 & 64.17 & 72.73 & 62.09 & 71.61 & 68.42 & 55.96 & 78.77 & 69.39 \\
Qwen3-235B-A22B-Instruct-2507 & 62.96 & 51.69 & 58.34 & 31.71 & 67.91 & 77.27 & 61.65 & 69.75 & 51.76 & 47.18 & 70.99 & 59.18 & 73.14 & 67.76 & 75.12 & 73.49 & 75.82 & 77.17 & 61.62 & 77.54 & 71.43 \\
deepseek-v3 & 66.67 & 38.2 & 46.09 & 31.71 & 63.26 & 81.82 & 61.7 & 65.55 & 41.18 & 49.01 & 61.11 & 46.94 & 60.71 & 44.89 & 62.2 & 55.58 & 61.98 & 61.92 & 45.45 & 66.15 & 51.7 \\
gpt-4o & 62.96 & 47.19 & 59.95 & 36.59 & 63.72 & 100 & 75.87 & 54.82 & 63.87 & 50.59 & 70.22 & 57.14 & 74.06 & 68.6 & 73.21 & 69.07 & 74.96 & 77.19 & 62.22 & 74.46 & 61.9 \\
\bottomrule
\end{tabular}
}
\end{table}

\subsection{Zero-Shot sub-domain wise Performance Across BhashaBench V1 Domains}

\begin{small}
\begin{longtable}{p{4cm}*{8}{c}}
\caption{Performance of GEMMA model family across sub-domains in BhashaBench v1, comparing base and instruction-tuned variants of different model sizes (270M, 2B, 9B, 27B)}
\label{tab:model_performance} \\
\toprule
\textbf{Subject Domain} & \textbf{270m} & \textbf{270m-it} & \textbf{2b} & \textbf{2b-it} & \textbf{9b} & \textbf{9b-it} & \textbf{27b} & \textbf{27b-it} \\
\midrule
\endfirsthead

\multicolumn{9}{c}{\tablename\ \thetable\ -- \textit{Continued from previous page}} \\
\toprule
\textbf{Subject Domain} & \textbf{270m} & \textbf{270m-it} & \textbf{2b} & \textbf{2b-it} & \textbf{9b} & \textbf{9b-it} & \textbf{27b} & \textbf{27b-it} \\
\midrule
\endhead

\midrule
\multicolumn{9}{r}{\textit{Continued on next page}} \\
\endfoot

\bottomrule
\endlastfoot

\multicolumn{9}{c}{\textbf{BBA}} \\
\midrule
Administration, AYUSH \& Miscellaneous & 34.45 & 28.57 & 40.34 & 34.45 & 63.03 & 51.26 & 60.5 & 57.14 \\
Agad Tantra \& Forensic Medicine & 25.89 & 27.94 & 31.18 & 27.94 & 48.21 & 39.35 & 49.4 & 42.25 \\
Ayurvedic Literature \& History & 26.96 & 23.53 & 31.37 & 28.92 & 46.08 & 31.86 & 43.14 & 42.16 \\
Dravyaguna \& Bhaishajya & 28.4 & 26.35 & 30.08 & 27.79 & 38.43 & 32.74 & 39.64 & 33.68 \\
Kaumarbhritya \& Pediatrics & 28.57 & 27.03 & 38.8 & 28.15 & 46.22 & 31.65 & 47.9 & 36.55 \\
Kayachikitsa (General Medicine \& Internal Medicine in Ayurveda) & 29.45 & 25.72 & 36.76 & 29.1 & 47.16 & 34.3 & 50.8 & 36.89 \\
Panchakarma \& Rasayana & 26.83 & 23.7 & 30.2 & 26.53 & 32.49 & 28.36 & 37.84 & 33.94 \\
Research \& Statistics & 27.14 & 25.24 & 60 & 34.29 & 77.62 & 53.81 & 78.1 & 57.62 \\
Roga Vigyana (Diagnostics \& Pathology) & 31.25 & 38.75 & 45 & 35 & 65 & 55 & 72.5 & 56.25 \\
Samhita \& Siddhanta (Fundamentals) & 30.89 & 29.07 & 33.29 & 28.42 & 37.7 & 30.95 & 43.93 & 34.59 \\
Shalakya Tantra (ENT, Eye, Dentistry) & 25.89 & 21.93 & 34.74 & 21.66 & 44.69 & 31.2 & 45.78 & 34.88 \\
Shalya Tantra (Surgery) & 26.0 & 23 & 31.94 & 26.05 & 45.06 & 31.75 & 44.87 & 39.16 \\
Sharir (Anatomy \& Physiology) & 24.59 & 26.45 & 33.28 & 27.79 & 46.95 & 34.75 & 51.04 & 40.19 \\
Stri Roga \& Prasuti Tantra (Gynecology \& Obstetrics) & 24.68 & 24.09 & 34.59 & 29.87 & 46.99 & 40.73 & 53.96 & 42.38 \\
Swasthavritta \& Public Health & 34.88 & 30.24 & 49.67 & 39.07 & 67.33 & 49.01 & 71.52 & 59.82 \\
Yoga \& Psychology & 30.85 & 26.6 & 43.62 & 32.98 & 57.45 & 37.77 & 61.7 & 46.81 \\
\midrule
\multicolumn{9}{c}{\textbf{BBF}} \\
\midrule
Accounting & 26.78 & 26 & 31.31 & 30.53 & 41.14 & 38.03 & 44.11 & 39.46 \\
Banking Services & 23.4 & 25.19 & 37.75 & 34.67 & 53.8 & 47.82 & 60.8 & 54.06 \\
Behavioral Finance & 31.34 & 28.36 & 47.76 & 46.27 & 50.75 & 59.7 & 52.24 & 52.24 \\
Business Management & 26.51 & 25.3 & 55.42 & 45.78 & 63.86 & 50.6 & 75.9 & 62.65 \\
Commerce & 28.04 & 22.48 & 32.79 & 31.05 & 40.32 & 39.17 & 48.78 & 41.25 \\
Corporate Finance \& Investment & 25.16 & 23.52 & 31.1 & 31.98 & 44.4 & 39.56 & 50.55 & 43.19 \\
Data \& Analytics in Finance & 23.62 & 24.41 & 32.28 & 27.56 & 38.58 & 30.71 & 44.88 & 29.13 \\
Economics \& Development Studies & 22.99 & 20.8 & 37.96 & 41.24 & 62.41 & 45.62 & 63.87 & 46.72 \\
Energy, Infrastructure \& Finance & 20.73 & 31.71 & 34.15 & 28.05 & 43.9 & 50 & 51.22 & 42.68 \\
Environmental Finance & 22.02 & 23.21 & 41.07 & 34.5 & 50 & 43.45 & 61.9 & 54.76 \\
Finance Education & 26.27 & 27.12 & 43.22 & 39.83 & 49.15 & 44.07 & 55.08 & 49.15 \\
Financial Markets & 31.91 & 25.53 & 53.19 & 36.17 & 51.06 & 44.68 & 63.83 & 55.32 \\
Financial Technology & 34.78 & 26.09 & 26.09 & 47.83 & 60.87 & 47.83 & 60.87 & 47.83 \\
General Knowledge & 24.3 & 26.35 & 41.37 & 38.4 & 57.7 & 51.02 & 61.78 & 52.5 \\
Governance \& Policy & 26.69 & 24.72 & 36.18 & 34.21 & 52.07 & 46.52 & 60.9 & 51.13 \\
Healthcare Economics & 27.19 & 30.7 & 40.35 & 39.47 & 57.89 & 50 & 61.4 & 51.75 \\
History, Sociology \& Cultural Studies of Finance & 18.11 & 25.98 & 40.94 & 41.73 & 60.63 & 51.18 & 64.57 & 57.48 \\
Information Technology Finance & 23.06 & 28.57 & 55.31 & 44.49 & 80 & 63.47 & 83.27 & 67.14 \\
Insurance \& Risk Management & 16.67 & 33.33 & 38.1 & 30.95 & 50 & 38.1 & 50 & 40.48 \\
Interdisciplinary Finance & 25.49 & 20.92 & 35.95 & 36.6 & 56.86 & 49.02 & 62.75 & 51.63 \\
International Finance \& Trade & 21.69 & 16.87 & 42.17 & 42.17 & 66.27 & 59.04 & 73.49 & 61.45 \\
Language \& Communication & 22.73 & 23.04 & 39.43 & 40.06 & 59.83 & 47.89 & 61.1 & 49.79 \\
Legal Finance & 32.35 & 29.41 & 35.29 & 41.18 & 47.06 & 35.29 & 50 & 50 \\
Marketing Finance & 26.19 & 26.19 & 47.62 & 35.71 & 76.19 & 61.9 & 66.67 & 59.52 \\
Mathematics for Finance & 24.83 & 23.76 & 28.96 & 25.96 & 33.81 & 31 & 38.53 & 32.69 \\
Problem Solving & 25.08 & 23.11 & 26.28 & 24.76 & 28.14 & 26.73 & 31.6 & 30.99 \\
Rural Economics & 25.67 & 29.89 & 39.46 & 40.61 & 57.47 & 50.19 & 68.2 & 54.79 \\
Science and Technology in Finance & 26.73 & 19.8 & 31.68 & 37.62 & 48.51 & 50.5 & 61.39 & 54.46 \\
Sports, Media \& Finance Linkages & 15.56 & 20 & 37.78 & 48.89 & 62.22 & 62.22 & 66.67 & 64.44 \\
Taxation \& Regulatory Compliance & 32.26 & 26.45 & 36.13 & 45.81 & 58.71 & 51.61 & 64.52 & 52.9 \\
\midrule
\multicolumn{9}{c}{\textbf{BBK}} \\
\midrule
Agri-Environmental \& Allied Disciplines & 26.14 & 26.7 & 29.55 & 36.93 & 48.86 & 46.02 & 48.86 & 54.55 \\
Agricultural Biotechnology & 26.15 & 29.77 & 54.2 & 43.13 & 75.19 & 63.93 & 77.67 & 70.61 \\
Agricultural Chemistry \& Biochemistry & 23.84 & 24.2 & 40.93 & 33.1 & 54.8 & 51.25 & 61.92 & 56.23 \\
Agricultural Economics \& Policy & 28.55 & 25.36 & 43.06 & 38.76 & 56.3 & 49.6 & 62.2 & 54.39 \\
Agricultural Engineering \& Technology & 29.51 & 25 & 38.93 & 26.64 & 50.41 & 34.02 & 58.61 & 41.8 \\
Agricultural Extension Education & 27.13 & 28.68 & 37.47 & 34.75 & 53.75 & 49.74 & 60.47 & 55.04 \\
Agricultural Microbiology & 21.62 & 25.23 & 48.65 & 35.14 & 69.37 & 49.55 & 75.68 & 64.86 \\
Agriculture Communication & 22.83 & 22.44 & 38.19 & 33.86 & 55.91 & 50.39 & 64.57 & 53.15 \\
Agriculture Information Technology & 27.89 & 28.42 & 39.47 & 43.16 & 57.89 & 55.79 & 61.05 & 59.47 \\
Agronomy & 26.47 & 26.84 & 38.64 & 33.56 & 52.44 & 45.45 & 57.33 & 50.32 \\
Animal Sciences & 31.08 & 24.32 & 52.7 & 43.24 & 64.19 & 50.68 & 66.22 & 55.41 \\
Crop Sciences & 24.95 & 27.69 & 38.43 & 37.34 & 46.45 & 48.09 & 51.73 & 51.73 \\
Dairy \& Poultry Science & 34.83 & 24.72 & 46.07 & 32.58 & 57.3 & 46.07 & 66.29 & 53.93 \\
Entomology & 27.16 & 26.87 & 38.36 & 34.63 & 57.04 & 50.14 & 61.21 & 55.32 \\
Fisheries and Aquaculture & 32.35 & 11.76 & 35.29 & 38.24 & 58.82 & 47.06 & 73.53 & 50 \\
General Knowledge \& Reasoning & 26.32 & 27.99 & 39.18 & 32.83 & 51.89 & 48.41 & 56.58 & 52.5 \\
Genetics and Plant Breeding & 25.96 & 27.51 & 39.85 & 36.25 & 51.93 & 52.96 & 58.61 & 55.01 \\
Horticulture & 25.56 & 26.18 & 36.28 & 32.42 & 48.65 & 41.21 & 53.67 & 48.12 \\
Natural Resource Management & 27.98 & 28.5 & 38.34 & 33.68 & 48.7 & 47.67 & 52.33 & 50.26 \\
Nematology & 26.09 & 31.52 & 28.8 & 32.07 & 40.76 & 40.22 & 48.91 & 48.37 \\
Plant Pathology & 23.17 & 27.71 & 36.27 & 34.51 & 53.65 & 47.36 & 55.67 & 54.91 \\
Plant Sciences \& Physiology & 28.68 & 26.36 & 45.74 & 29.46 & 67.44 & 51.94 & 71.32 & 55.81 \\
Seed Science and Technology & 22.28 & 33.66 & 35.64 & 32.18 & 45.05 & 43.56 & 47.52 & 50.5 \\
Soil Science & 25.0 & 28 & 35 & 35.08 & 52.17 & 43.63 & 56.6 & 53.87 \\
Veterinary Sciences & 39.58 & 29.17 & 60.42 & 35.42 & 83.33 & 66.67 & 85.42 & 77.08 \\
\midrule
\multicolumn{9}{c}{\textbf{BBL}} \\
\midrule
Civil Litigation \& Procedure & 25.26 & 27.36 & 33.6 & 32.33 & 49.61 & 40.2 & 57.91 & 43.92 \\
Constitutional \& Administrative Law & 25.27 & 25.57 & 37.55 & 33.75 & 58.94 & 46.08 & 65.31 & 52.84 \\
Consumer \& Competition Law & 32 & 25.33 & 33.33 & 37.33 & 57.33 & 53.33 & 69.33 & 61.33 \\
Corporate \& Commercial Law & 25.33 & 25.15 & 36.48 & 31.0 & 53 & 39.81 & 60.04 & 45.59 \\
Criminal Law \& Justice & 25.57 & 25.75 & 31.67 & 32.47 & 50.31 & 42.9 & 57.39 & 45.97 \\
Employment \& Labour Law & 24.57 & 29.71 & 33.14 & 37.14 & 54.29 & 44.57 & 60.57 & 46.86 \\
Environmental \& Energy Law & 21.63 & 22.56 & 34.19 & 32.33 & 53.26 & 41.4 & 61.16 & 49.77 \\
Family \& Personal Law & 25.83 & 26.34 & 33.91 & 31.18 & 47.83 & 37.74 & 57.62 & 44.2 \\
General Academic Subjects & 29.27 & 25.97 & 44.99 & 38.84 & 67.94 & 53.76 & 73.52 & 59.68 \\
Healthcare \& Medical Law & 32 & 32 & 52 & 40 & 72 & 52 & 76 & 72 \\
Human Rights \& Social Justice & 5.26 & 10.53 & 47.37 & 15.79 & 47.37 & 26.32 & 42.11 & 31.58 \\
Intellectual Property Law & 25.27 & 27.47 & 54.95 & 48.35 & 72.53 & 56.04 & 70.33 & 59.34 \\
Interdisciplinary Studies & 20.39 & 26.72 & 39.67 & 37.19 & 61.98 & 49.86 & 70.8 & 57.58 \\
International \& Comparative Law & 24.22 & 23.91 & 44.28 & 37.32 & 65.49 & 52.18 & 70.17 & 58.84 \\
Legal Skills \& Communication & 27.7 & 23.28 & 25.61 & 27.94 & 36.76 & 32.35 & 39.46 & 36.52 \\
Legal Theory \& Jurisprudence & 25.4 & 27.59 & 38.21 & 35.33 & 57.49 & 48.06 & 64.6 & 51.23 \\
Media \& Entertainment Law & 16.67 & 33.33 & 35.19 & 44.44 & 61.11 & 51.85 & 72.22 & 66.67 \\
Real Estate \& Property Law & 24.8 & 22.8 & 31 & 28.3 & 47.54 & 34.34 & 53.42 & 38 \\
Tax \& Revenue Law & 23.81 & 26.41 & 38.1 & 32.03 & 51.52 & 38.1 & 65.37 & 48.05 \\
Technology \& Cyber Law & 28.46 & 28.46 & 47.15 & 44.72 & 64.23 & 59.35 & 75.61 & 69.92 \\
\end{longtable}
\end{small}

\begin{small}
\begin{longtable}{p{4.8cm}*{6}{c}}
\caption{Performance of Llama model family across sub-domains in BhashaBench v1, comparing base and instruction-tuned variants (1B, 3B, 8B)}
\label{tab:llama_performance} \\
\toprule
\textbf{Subject Domain} & \textbf{3.2-1B} & \textbf{3.2-1B-it} & \textbf{3.2-3B} & \textbf{3.2-3B-it} & \textbf{3.1-8B} & \textbf{3.1-8B-it} \\
\midrule
\endfirsthead
\multicolumn{7}{c}{\tablename\ \thetable\ -- \textit{Continued from previous page}} \\
\toprule
\textbf{Subject Domain} & \textbf{3.2-1B} & \textbf{3.2-1B-it} & \textbf{3.2-3B} & \textbf{3.2-3B-it} & \textbf{3.1-8B} & \textbf{3.1-8B-it} \\
\midrule
\endhead
\midrule
\multicolumn{7}{r}{\textit{Continued on next page}} \\
\endfoot
\bottomrule
\endlastfoot
\multicolumn{7}{c}{\textbf{BBA}} \\
\midrule
Administration, AYUSH \& Miscellaneous & 36.97 & 35.29 & 31.93 & 39.5 & 41.18 & 44.54 \\
Agad Tantra \& Forensic Medicine & 28.28 & 27.09 & 35.09 & 39.01 & 33.9 & 35.6 \\
Ayurvedic Literature \& History & 27.45 & 30.88 & 29.9 & 33.33 & 30.88 & 36.27 \\
Dravyaguna \& Bhaishajya & 26.58 & 26.92 & 26.95 & 30.11 & 29.24 & 31.53 \\
Kaumarbhritya \& Pediatrics & 28.57 & 25.63 & 29.41 & 32.91 & 31.09 & 35.71 \\
Kayachikitsa (General Medicine \& Internal Medicine in Ayurveda) & 29.04 & 24.92 & 31.33 & 34.84 & 34.24 & 34.78 \\
Panchakarma \& Rasayana & 27.06 & 25.76 & 27.06 & 30.2 & 29.05 & 28.75 \\
Research \& Statistics & 27.14 & 29.5 & 40 & 44.29 & 47.62 & 54.76 \\
Roga Vigyana (Diagnostics \& Pathology) & 35 & 25 & 45 & 42.5 & 50 & 61.25 \\
Samhita \& Siddhanta (Fundamentals) & 29.92 & 26.15 & 31.28 & 27.84 & 33.55 & 27.9 \\
Shalakya Tantra (ENT, Eye, Dentistry) & 27.25 & 26.84 & 29.43 & 35.29 & 31.61 & 37.47 \\
Shalya Tantra (Surgery) & 25.48 & 25.48 & 28.33 & 30.8 & 35.17 & 34.6 \\
Sharir (Anatomy \& Physiology) & 27.12 & 25.19 & 29.49 & 33.66 & 32.76 & 38.93 \\
Stri Roga \& Prasuti Tantra (Gynecology \& Obstetrics) & 27.27 & 28.1 & 31.88 & 33.6 & 34 & 36.36 \\
Swasthavritta \& Public Health & 34 & 32.67 & 40.62 & 51.21 & 47.46 & 57.17 \\
Yoga \& Psychology & 26.6 & 24.47 & 32.45 & 31.38 & 43.62 & 34.57 \\
\midrule
\multicolumn{7}{c}{\textbf{BBF}} \\
\midrule
Accounting & 27.3 & 26.13 & 30.66 & 27.68 & 34.54 & 30.66 \\
Banking Services & 30.49 & 28.18 & 38.34 & 38.68 & 40.48 & 42.36 \\
Behavioral Finance & 37.31 & 28.36 & 35.82 & 37.31 & 47.76 & 49.25 \\
Business Management & 26.51 & 26.51 & 43.37 & 53.01 & 50.6 & 60.24 \\
Commerce & 28.51 & 27.46 & 32.1 & 31.52 & 34.41 & 31.98 \\
Corporate Finance \& Investment & 27.58 & 26.37 & 29.56 & 35.05 & 37.91 & 39.23 \\
Data \& Analytics in Finance & 22.83 & 18.11 & 31.5 & 20.47 & 32.28 & 31.5 \\
Economics \& Development Studies & 29.56 & 32.85 & 36.13 & 40.51 & 39.42 & 48.18 \\
Energy, Infrastructure \& Finance & 29.27 & 28.05 & 32.93 & 39.02 & 42.68 & 40.24 \\
Environmental Finance & 25 & 29.76 & 39.29 & 38.69 & 41.07 & 51.19 \\
Finance Education & 29.66 & 25.42 & 49.15 & 34.75 & 44.92 & 47.46 \\
Financial Markets & 36.17 & 29.79 & 57.45 & 48.94 & 40.43 & 51.06 \\
Financial Technology & 17.39 & 13.04 & 21.74 & 34.78 & 43.48 & 47.83 \\
General Knowledge & 31.35 & 28.94 & 37.48 & 43.04 & 42.3 & 50.09 \\
Governance \& Policy & 28.76 & 27.63 & 34.3 & 39.29 & 40.13 & 47.84 \\
Healthcare Economics & 31.58 & 31.58 & 38.6 & 41.23 & 50.88 & 51.75 \\
History, Sociology \& Cultural Studies of Finance & 24.41 & 30.71 & 37.01 & 44.88 & 41.73 & 61.42 \\
Information Technology Finance & 31.63 & 35.51 & 46.33 & 53.06 & 59.59 & 66.33 \\
Insurance \& Risk Management & 19.05 & 26.19 & 30.95 & 38.1 & 42.86 & 40.48 \\
Interdisciplinary Finance & 26.14 & 30.72 & 37.25 & 33.33 & 37.91 & 54.9 \\
International Finance \& Trade & 27.71 & 34.94 & 36.14 & 39.76 & 45.78 & 54.22 \\
Language \& Communication & 32.45 & 29.18 & 35.62 & 40.59 & 42.49 & 43.66 \\
Legal Finance & 26.47 & 20.59 & 29.41 & 20.59 & 38.24 & 35.29 \\
Marketing Finance & 23.81 & 38.1 & 38.1 & 38.1 & 59.52 & 52.38 \\
Mathematics for Finance & 27.31 & 24.91 & 28.96 & 27.57 & 29.97 & 26.3 \\
Problem Solving & 24.67 & 23.65 & 27.08 & 25.15 & 28.1 & 24.6 \\
Rural Economics & 27.97 & 30.65 & 33.33 & 44.83 & 42.53 & 51.72 \\
Science and Technology in Finance & 21.78 & 30.69 & 31.68 & 41.58 & 38.61 & 35.64 \\
Sports, Media \& Finance Linkages & 33.33 & 28.89 & 48.89 & 42.22 & 51.11 & 48.89 \\
Taxation \& Regulatory Compliance & 36.13 & 31.61 & 43.87 & 47.1 & 47.1 & 50.97 \\
\midrule
\multicolumn{7}{c}{\textbf{BBK}} \\
\midrule
Agri-Environmental \& Allied Disciplines & 31.82 & 32.95 & 25 & 36.36 & 30.68 & 47.73 \\
Agricultural Biotechnology & 31.11 & 28.63 & 34.35 & 50.95 & 48.85 & 58.78 \\
Agricultural Chemistry \& Biochemistry & 27.05 & 22.78 & 31.32 & 33.81 & 38.79 & 48.75 \\
Agricultural Economics \& Policy & 29.98 & 25.52 & 35.09 & 38.12 & 40.35 & 46.73 \\
Agricultural Engineering \& Technology & 27.46 & 26.23 & 32.79 & 33.2 & 38.93 & 41.8 \\
Agricultural Extension Education & 30.88 & 29.46 & 32.3 & 41.99 & 40.31 & 48.19 \\
Agricultural Microbiology & 34.23 & 36.04 & 31.53 & 53.15 & 38.74 & 54.95 \\
Agriculture Communication & 33.07 & 28.35 & 29.53 & 44.49 & 36.61 & 49.21 \\
Agriculture Information Technology & 30.53 & 31.58 & 44.21 & 45.79 & 46.32 & 45.79 \\
Agronomy & 27.92 & 28.77 & 31.84 & 37.22 & 37.2 & 43.34 \\
Animal Sciences & 25.68 & 34.46 & 36.49 & 41.89 & 46.62 & 45.95 \\
Crop Sciences & 31.15 & 26.41 & 29.87 & 35.34 & 38.25 & 40.8 \\
Dairy \& Poultry Science & 35.96 & 31.46 & 30.34 & 37.08 & 41.57 & 44.94 \\
Entomology & 29.02 & 27.59 & 35.49 & 35.49 & 38.79 & 47.7 \\
Fisheries and Aquaculture & 29.41 & 41.18 & 38.24 & 55.88 & 38.24 & 52.94 \\
General Knowledge \& Reasoning & 28.44 & 27.53 & 33.13 & 39.64 & 38.88 & 42.66 \\
Genetics and Plant Breeding & 30.59 & 30.08 & 28.02 & 38.3 & 40.62 & 43.19 \\
Horticulture & 27.05 & 28.6 & 31.21 & 36.86 & 35.89 & 43 \\
Natural Resource Management & 28.5 & 26.42 & 29.02 & 37.82 & 33.16 & 44.56 \\
Nematology & 22.83 & 28.26 & 28.26 & 29.35 & 35.33 & 41.3 \\
Plant Pathology & 28.97 & 30.48 & 27.96 & 42.82 & 34.01 & 44.84 \\
Plant Sciences \& Physiology & 28.68 & 31.78 & 37.98 & 50.39 & 43.41 & 54.26 \\
Seed Science and Technology & 29.7 & 28.71 & 27.72 & 37.13 & 35.15 & 38.61 \\
Soil Science & 31.25 & 29.92 & 31.69 & 38.84 & 37.14 & 45.25 \\
Veterinary Sciences & 27.08 & 14.58 & 37.5 & 47.92 & 43.75 & 70.83 \\
\midrule
\multicolumn{7}{c}{\textbf{BBL}} \\
\midrule
Civil Litigation \& Procedure & 29.32 & 28.18 & 32.4 & 34.97 & 36.68 & 42.66 \\
Constitutional \& Administrative Law & 29.54 & 28.15 & 36.22 & 40.62 & 42.28 & 49.46 \\
Consumer \& Competition Law & 28 & 22.67 & 28 & 34.67 & 46.67 & 41.33 \\
Corporate \& Commercial Law & 27.7 & 28.63 & 29.78 & 34.67 & 35.15 & 42.67 \\
Criminal Law \& Justice & 27.09 & 26.98 & 30.01 & 33.66 & 35.21 & 42.72 \\
Employment \& Labour Law & 23.43 & 25.71 & 28.57 & 29.1 & 32 & 40 \\
Environmental \& Energy Law & 27.67 & 24.42 & 33.49 & 37.91 & 39.07 & 45.81 \\
Family \& Personal Law & 24.12 & 28.86 & 29.06 & 31.69 & 34.21 & 39.86 \\
General Academic Subjects & 29.21 & 32.52 & 37.47 & 43.91 & 46.87 & 52.68 \\
Healthcare \& Medical Law & 40 & 20 & 68 & 40 & 64 & 60 \\
Human Rights \& Social Justice & 21.05 & 42.11 & 36.84 & 26.32 & 31.58 & 36.84 \\
Intellectual Property Law & 30.77 & 31.87 & 46.15 & 45.05 & 56.04 & 58.24 \\
Interdisciplinary Studies & 33.33 & 28.1 & 38.57 & 41.32 & 43.25 & 53.72 \\
International \& Comparative Law & 30.87 & 30.35 & 40.02 & 45.22 & 46.88 & 54.47 \\
Legal Skills \& Communication & 25.74 & 27.33 & 28.68 & 30.15 & 28.31 & 32.72 \\
Legal Theory \& Jurisprudence & 29.63 & 28.36 & 33.92 & 39.69 & 41.66 & 46.87 \\
Media \& Entertainment Law & 33.33 & 35.19 & 42.59 & 51.85 & 38.89 & 53.7 \\
Real Estate \& Property Law & 23.53 & 25.91 & 29.89 & 31.96 & 31.48 & 38.16 \\
Tax \& Revenue Law & 27.71 & 31.6 & 40.26 & 38.1 & 41.56 & 43.29 \\
Technology \& Cyber Law & 30.89 & 41.46 & 48.78 & 49.59 & 51.22 & 60.16 \\
\end{longtable}
\end{small}

\begin{small}
\begin{longtable}{p{7cm}*{3}{c}}
\caption{Performance of Qwen model family across sub-domains in BhashaBench v1, comparing base and instruction-tuned variants (3B, 235B)}
\label{tab:qwen_performance} \\
\toprule
\textbf{Subject Domain} & \textbf{2.5-3B} & \textbf{2.5-3B-it} & \textbf{3-235B-A22B-it-2507} \\
\midrule
\endfirsthead
\multicolumn{4}{c}{\tablename\ \thetable\ -- \textit{Continued from previous page}} \\
\toprule
\textbf{Subject Domain} & \textbf{2.5-3B} & \textbf{2.5-3B-it} & \textbf{3-235B-A22B-it-2507} \\
\midrule
\endhead
\midrule
\multicolumn{4}{r}{\textit{Continued on next page}} \\
\endfoot
\bottomrule
\endlastfoot
\multicolumn{4}{c}{\textbf{BBA}} \\
\midrule
Administration, AYUSH \& Miscellaneous & 47.06 & 38.66 & 73.11 \\
Agad Tantra \& Forensic Medicine & 39.86 & 32.71 & 63.88 \\
Ayurvedic Literature \& History & 38.73 & 29.9 & 55.88 \\
Dravyaguna \& Bhaishajya & 32.57 & 28.94 & 49.43 \\
Kaumarbhritya \& Pediatrics & 38.52 & 30.11 & 55.32 \\
Kayachikitsa (General Medicine \& Internal Medicine in Ayurveda) & 38.61 & 35.07 & 59.48 \\
Panchakarma \& Rasayana & 30.35 & 29.59 & 49.54 \\
Research \& Statistics & 62.86 & 52.86 & 91.43 \\
Roga Vigyana (Diagnostics \& Pathology) & 58.75 & 53.75 & 82.5 \\
Samhita \& Siddhanta (Fundamentals) & 36.79 & 31.93 & 55.22 \\
Shalakya Tantra (ENT, Eye, Dentistry) & 35.56 & 31.74 & 59.67 \\
Shalya Tantra (Surgery) & 37.45 & 33.08 & 60.46 \\
Sharir (Anatomy \& Physiology) & 37.44 & 31.35 & 60.1 \\
Stri Roga \& Prasuti Tantra (Gynecology \& Obstetrics) & 40.73 & 34.24 & 66.82 \\
Swasthavritta \& Public Health & 50.99 & 43.49 & 82.56 \\
Yoga \& Psychology & 44.68 & 36.17 & 75.53 \\
\midrule
\multicolumn{4}{c}{\textbf{BBF}} \\
\midrule
Accounting & 38.94 & 31.82 & 63.52 \\
Banking Services & 43.3 & 36.89 & 71.22 \\
Behavioral Finance & 52.24 & 44.78 & 71.64 \\
Business Management & 60.24 & 40.96 & 84.34 \\
Commerce & 43.57 & 33.72 & 63.62 \\
Corporate Finance \& Investment & 40.22 & 37.58 & 63.52 \\
Data \& Analytics in Finance & 35.43 & 28.35 & 53.54 \\
Economics \& Development Studies & 43.8 & 44.16 & 73.36 \\
Energy, Infrastructure \& Finance & 45.12 & 30.49 & 71.95 \\
Environmental Finance & 47.62 & 44.05 & 82.74 \\
Finance Education & 50.85 & 43.22 & 69.49 \\
Financial Markets & 42.55 & 42.55 & 70.21 \\
Financial Technology & 47.83 & 39.13 & 78.26 \\
General Knowledge & 41.56 & 38.22 & 74.95 \\
Governance \& Policy & 45.3 & 38.16 & 74.15 \\
Healthcare Economics & 48.25 & 45.61 & 78.95 \\
History, Sociology \& Cultural Studies of Finance & 38.58 & 38.58 & 83.46 \\
Information Technology Finance & 64.9 & 58.16 & 92.24 \\
Insurance \& Risk Management & 30.95 & 38.1 & 64.29 \\
Interdisciplinary Finance & 41.83 & 36.6 & 79.74 \\
International Finance \& Trade & 49.4 & 42.17 & 78.31 \\
Language \& Communication & 45.77 & 42.71 & 77.06 \\
Legal Finance & 38.24 & 23.53 & 76.47 \\
Marketing Finance & 69.05 & 50 & 85.71 \\
Mathematics for Finance & 34.18 & 29.85 & 58.04 \\
Problem Solving & 27.88 & 26.2 & 47.12 \\
Rural Economics & 47.13 & 45.21 & 80.46 \\
Science and Technology in Finance & 40.59 & 43.56 & 72.28 \\
Sports, Media \& Finance Linkages & 44.44 & 53.33 & 68.89 \\
Taxation \& Regulatory Compliance & 56.13 & 38.71 & 74.84 \\
\midrule
\multicolumn{4}{c}{\textbf{BBK}} \\
\midrule
Agri-Environmental \& Allied Disciplines & 43.75 & 43.18 & 75.57 \\
Agricultural Biotechnology & 55.34 & 51.15 & 91.6 \\
Agricultural Chemistry \& Biochemistry & 44.48 & 38.43 & 83.63 \\
Agricultural Economics \& Policy & 46.41 & 43.38 & 73.21 \\
Agricultural Engineering \& Technology & 41.39 & 37.3 & 67.21 \\
Agricultural Extension Education & 46.25 & 42.51 & 72.87 \\
Agricultural Microbiology & 54.05 & 43.24 & 90.99 \\
Agriculture Communication & 44.49 & 44.49 & 78.35 \\
Agriculture Information Technology & 52.63 & 54.21 & 74.74 \\
Agronomy & 41.73 & 38.89 & 71.92 \\
Animal Sciences & 47.97 & 46.62 & 77.7 \\
Crop Sciences & 42.08 & 36.79 & 67.4 \\
Dairy \& Poultry Science & 52.81 & 46.07 & 75.28 \\
Entomology & 39.94 & 39.66 & 77.44 \\
Fisheries and Aquaculture & 38.24 & 50 & 79.41 \\
General Knowledge \& Reasoning & 44.48 & 41.6 & 73.22 \\
Genetics and Plant Breeding & 43.44 & 44.22 & 76.86 \\
Horticulture & 37.25 & 35.41 & 64.98 \\
Natural Resource Management & 37.82 & 37.31 & 65.8 \\
Nematology & 33.15 & 39.13 & 63.04 \\
Plant Pathology & 40.55 & 36.52 & 78.34 \\
Plant Sciences \& Physiology & 45.74 & 48.06 & 86.82 \\
Seed Science and Technology & 42.08 & 34.65 & 66.34 \\
Soil Science & 42 & 39.35 & 72.37 \\
Veterinary Sciences & 45.83 & 50 & 87.5 \\
\midrule
\multicolumn{4}{c}{\textbf{BBL}} \\
\midrule
Civil Litigation \& Procedure & 38.65 & 35.31 & 72.12 \\
Constitutional \& Administrative Law & 43.67 & 37.93 & 82.65 \\
Consumer \& Competition Law & 36 & 46.67 & 82.67 \\
Corporate \& Commercial Law & 40.74 & 37.7 & 77.11 \\
Criminal Law \& Justice & 38.21 & 34.45 & 75.44 \\
Employment \& Labour Law & 39.43 & 37.14 & 71.43 \\
Environmental \& Energy Law & 44.65 & 38.84 & 76.74 \\
Family \& Personal Law & 38.35 & 32.8 & 74.37 \\
General Academic Subjects & 53.82 & 45.44 & 85.82 \\
Healthcare \& Medical Law & 56 & 40 & 88 \\
Human Rights \& Social Justice & 47.37 & 31.58 & 73.68 \\
Intellectual Property Law & 60.44 & 54.95 & 87.91 \\
Interdisciplinary Studies & 49.31 & 44.08 & 84.85 \\
International \& Comparative Law & 47.51 & 43.76 & 83.89 \\
Legal Skills \& Communication & 32.35 & 31.74 & 61.27 \\
Legal Theory \& Jurisprudence & 46.45 & 40.04 & 79.38 \\
Media \& Entertainment Law & 42.59 & 33.33 & 79.63 \\
Real Estate \& Property Law & 36.09 & 33.55 & 71.7 \\
Tax \& Revenue Law & 39.83 & 37.66 & 74.03 \\
Technology \& Cyber Law & 58.54 & 59.35 & 86.18 \\
\end{longtable}
\end{small}

\begin{small}
\begin{longtable}{p{8cm}*{3}{c}}
\caption{Performance of GPT model family across sub-domains in BhashaBench v1, comparing different model sizes (20B, 120B, GPT-4o)}
\label{tab:gpt_performance} \\
\toprule
\textbf{Subject Domain} & \textbf{gpt-oss-20b} & \textbf{gpt-oss-120b} & \textbf{gpt-4o} \\
\midrule
\endfirsthead
\multicolumn{4}{c}{\tablename\ \thetable\ -- \textit{Continued from previous page}} \\
\toprule
\textbf{Subject Domain} & \textbf{gpt-oss-20b} & \textbf{gpt-oss-120b} & \textbf{gpt-4o} \\
\midrule
\endhead
\midrule
\multicolumn{4}{r}{\textit{Continued on next page}} \\
\endfoot
\bottomrule
\endlastfoot
\multicolumn{4}{c}{\textbf{BBA}} \\
\midrule
Administration, AYUSH \& Miscellaneous & 53.78 & 79.83 & 75.63 \\
Agad Tantra \& Forensic Medicine & 39.52 & 60.14 & 63.54 \\
Ayurvedic Literature \& History & 33.82 & 51.47 & 59.31 \\
Dravyaguna \& Bhaishajya & 30.75 & 44.48 & 54.78 \\
Kaumarbhritya \& Pediatrics & 35.99 & 51.4 & 56.58 \\
Kayachikitsa (General Medicine \& Internal Medicine in Ayurveda) & 39.06 & 54.69 & 60.69 \\
Panchakarma \& Rasayana & 28.36 & 41.44 & 50.76 \\
Research \& Statistics & 70.95 & 86.67 & 90 \\
Roga Vigyana (Diagnostics \& Pathology) & 66.25 & 82.5 & 81.25 \\
Samhita \& Siddhanta (Fundamentals) & 30.63 & 46.07 & 53.41 \\
Shalakya Tantra (ENT, Eye, Dentistry) & 38.15 & 54.9 & 62.4 \\
Shalya Tantra (Surgery) & 35.36 & 55.13 & 61.41 \\
Sharir (Anatomy \& Physiology) & 39.75 & 57.06 & 62.7 \\
Stri Roga \& Prasuti Tantra (Gynecology \& Obstetrics) & 35.18 & 59.03 & 64.82 \\
Swasthavritta \& Public Health & 56.51 & 76.6 & 81.02 \\
Yoga \& Psychology & 41.49 & 70.74 & 73.94 \\
\midrule
\multicolumn{4}{c}{\textbf{BBF}} \\
\midrule
Accounting & 35.45 & 73.61 & 49.55 \\
Banking Services & 42.53 & 67.29 & 68.57 \\
Behavioral Finance & 50.75 & 77.61 & 76.12 \\
Business Management & 53.01 & 87.95 & 81.93 \\
Commerce & 37.89 & 69.76 & 54.46 \\
Corporate Finance \& Investment & 37.25 & 73.63 & 61.43 \\
Data \& Analytics in Finance & 34.65 & 51.97 & 44.09 \\
Economics \& Development Studies & 46.72 & 69.34 & 71.53 \\
Energy, Infrastructure \& Finance & 39.02 & 64.63 & 67.07 \\
Environmental Finance & 55.95 & 73.21 & 77.98 \\
Finance Education & 46.61 & 73.73 & 74.58 \\
Financial Markets & 61.7 & 59.57 & 72.34 \\
Financial Technology & 47.83 & 73.91 & 78.26 \\
General Knowledge & 48.42 & 77.18 & 77.18 \\
Governance \& Policy & 39.85 & 69.36 & 78.29 \\
Healthcare Economics & 49.12 & 78.07 & 80.7 \\
History, Sociology \& Cultural Studies of Finance & 48.03 & 68.5 & 87.4 \\
Information Technology Finance & 76.94 & 90.82 & 92.04 \\
Insurance \& Risk Management & 47.62 & 57.14 & 64.29 \\
Interdisciplinary Finance & 45.1 & 73.2 & 75.82 \\
International Finance \& Trade & 54.22 & 75.9 & 85.54 \\
Language \& Communication & 47.57 & 74.42 & 77.48 \\
Legal Finance & 41.18 & 64.71 & 76.47 \\
Marketing Finance & 61.9 & 85.71 & 78.57 \\
Mathematics for Finance & 30.05 & 76.16 & 41.28 \\
Problem Solving & 26.63 & 64.14 & 42.65 \\
Rural Economics & 47.89 & 75.86 & 82.76 \\
Science and Technology in Finance & 45.54 & 77.23 & 73.27 \\
Sports, Media \& Finance Linkages & 46.67 & 75.56 & 73.33 \\
Taxation \& Regulatory Compliance & 44.52 & 68.39 & 73.55 \\
\midrule
\multicolumn{4}{c}{\textbf{BBK}} \\
\midrule
Agri-Environmental \& Allied Disciplines & 41.48 & 73.86 & 74.43 \\
Agricultural Biotechnology & 65.27 & 89.69 & 89.31 \\
Agricultural Chemistry \& Biochemistry & 54.8 & 80.43 & 81.14 \\
Agricultural Economics \& Policy & 46.57 & 71.77 & 73.68 \\
Agricultural Engineering \& Technology & 39.75 & 62.7 & 66.8 \\
Agricultural Extension Education & 43.93 & 69.25 & 75.19 \\
Agricultural Microbiology & 53.15 & 89.19 & 94.59 \\
Agriculture Communication & 42.91 & 73.23 & 81.1 \\
Agriculture Information Technology & 51.58 & 75.26 & 68.42 \\
Agronomy & 44.1 & 68 & 72.43 \\
Animal Sciences & 53.38 & 69.59 & 76.35 \\
Crop Sciences & 41.71 & 64.66 & 68.85 \\
Dairy \& Poultry Science & 52.81 & 75.28 & 78.65 \\
Entomology & 48.28 & 72.84 & 77.87 \\
Fisheries and Aquaculture & 50 & 64.71 & 73.53 \\
General Knowledge \& Reasoning & 42.81 & 69.59 & 68.38 \\
Genetics and Plant Breeding & 44.47 & 74.04 & 75.84 \\
Horticulture & 41.26 & 61.88 & 70.14 \\
Natural Resource Management & 41.97 & 64.77 & 65.8 \\
Nematology & 42.93 & 64.13 & 64.67 \\
Plant Pathology & 41.56 & 71.03 & 78.34 \\
Plant Sciences \& Physiology & 51.94 & 82.17 & 88.37 \\
Seed Science and Technology & 35.15 & 64.85 & 65.84 \\
Soil Science & 42.45 & 70.67 & 73.18 \\
Veterinary Sciences & 56.25 & 87.5 & 93.75 \\
\midrule
\multicolumn{4}{c}{\textbf{BBL}} \\
\midrule
Civil Litigation \& Procedure & 34.63 & 59.01 & 71.91 \\
Constitutional \& Administrative Law & 41.06 & 75.56 & 83.15 \\
Consumer \& Competition Law & 33.33 & 72 & 81.33 \\
Corporate \& Commercial Law & 37.48 & 69.59 & 78.93 \\
Criminal Law \& Justice & 35.14 & 65.11 & 75.95 \\
Employment \& Labour Law & 33.14 & 62.86 & 73.14 \\
Environmental \& Energy Law & 41.4 & 69.3 & 73.26 \\
Family \& Personal Law & 37.03 & 63.87 & 72.86 \\
General Academic Subjects & 56.49 & 83.14 & 84.79 \\
Healthcare \& Medical Law & 60 & 92 & 92 \\
Human Rights \& Social Justice & 15.79 & 73.68 & 68.42 \\
Intellectual Property Law & 53.85 & 85.71 & 90.11 \\
Interdisciplinary Studies & 43.25 & 82.64 & 83.75 \\
International \& Comparative Law & 48.86 & 79.42 & 81.7 \\
Legal Skills \& Communication & 32.84 & 69.12 & 53.43 \\
Legal Theory \& Jurisprudence & 42.08 & 75.16 & 81.21 \\
Media \& Entertainment Law & 50 & 83.33 & 85.19 \\
Real Estate \& Property Law & 32.59 & 59.62 & 71.7 \\
Tax \& Revenue Law & 42.86 & 67.53 & 69.26 \\
Technology \& Cyber Law & 56.91 & 86.18 & 86.99 \\
\end{longtable}
\end{small}

\end{document}